\documentclass{article}

\usepackage[preprint]{neurips_2026}

\usepackage[utf8]{inputenc} 
\usepackage[T1]{fontenc}    
\usepackage{hyperref}       
\usepackage{url}            
\usepackage{booktabs}       
\usepackage{amsfonts}       
\usepackage{nicefrac}       
\usepackage{microtype}      
\usepackage{xcolor}         
\usepackage{graphicx}
\usepackage{xparse}
\usepackage{xcolor}
\definecolor{navyblue}{RGB}{62, 117, 176}
\definecolor{brickred}{RGB}{192, 70, 87}
\usepackage{subfig}
\usepackage{svg}
\DeclareUnicodeCharacter{2212}{-} 
\DeclareUnicodeCharacter{0394}{$\Delta$}
\usepackage{array} 
\usepackage{makecell}  
\NewDocumentCommand{\DeclareMathCommand}{mmm}{%
  \expandafter\NewDocumentCommand\csname #1\endcsname{#2}{\ensuremath{#3}}%
}

\NewDocumentCommand{\DeclareVar}{mm}{%
  \expandafter\NewDocumentCommand\csname #1\endcsname{o}{%
    \ensuremath{#2\IfValueT{##1}{_{##1}}}%
  }%
}

\usepackage{enumitem}
\setitemize{noitemsep,topsep=0pt,parsep=0pt,partopsep=0pt}

\newcommand{\transl}[2]{#1$^{(\text{#2})}$}

\DeclareVar{embs}{\mathit{X}}
\DeclareVar{data}{\mathit{D}}
\DeclareVar{mixer}{\mathit A}
\DeclareVar{unmixer}{\mathit A*}
\DeclareVar{xpca}{\mathit Z}
\DeclareVar{xica}{\mathit S}
\DeclareVar{xsca}{\mathit T}
\DeclareVar{R}{\mathbb R}
\DeclareMathCommand{Rn}{m}{\mathbb R^{#1}}
\DeclareMathCommand{Rnxn}{mm}{\mathbb R^{#1 \times #2}}
\DeclareVar{vocab}{\mathcal V}

\newcommand{\absvalue}[1]{\left\vert #1 \right\vert}

\newcommand{\quot}[1]{``#1''}

\newcommand{\rhref}[2]{\href{#2}{#1}}

\newcommand{\NN}{\text{NN}}

\newenvironment{tightquote}{%
  \list{}{\topsep=0pt\partopsep=0pt\listparindent=0pt%
          \itemindent=0pt\leftmargin=2em\rightmargin=2em}%
  \item\relax
}{%
  \endlist
}

\usepackage{todonotes}

\usepackage{amsmath} 
\usepackage{amssymb}
\usepackage{url}
\usepackage{hyperref}
\setcitestyle{authoryear,round}

\makeatletter
\newif\ifshowacks
\if@anonymous\showacksfalse\else\showackstrue\fi
\makeatother

\makeatletter
\newif\ifusepng
\iffalse\usepngtrue\else\usepngfalse\fi   
\makeatother

\title{Structure Retention in Embedding Spaces as a Predictor of Benchmark Performance}

\author{%
  Amanda Myntti$^1$ \\
  \texttt{aaamyn@utu.fi} \\
  \And
  Jenna Kanerva$^1$ \\
  \texttt{jmnybl@utu.fi}\\
  \And
  Veronika Laippala$^1$ \\
  \texttt{mavela@utu.fi}\\
  \And
  Filip Ginter$^{1,2}$ \\
  \texttt{figint@utu.fi}\\
  \AND 
  $^1$TurkuNLP, University of Turku, Finland; $^2$ELLIS Institute Finland \\
}

\begin{document}

\maketitle

\begin{abstract}
In this paper, we show that high-performing embedding models organize their embedding spaces in a consistent way. We evaluate 25 contemporary embedding models on five MTEB tasks spanning four diverse task categories (retrieval, bitext mining, pair classification, and summarization) in both English and multilingual settings, and reveal that nearest-neighbor overlap and magnitude differences in independent component analysis (ICA) between paired text instances strongly correlate (even up to 0.97) with performance on the given task.  Ultimately, we show that embedding tasks display varying degrees of linearity and reliance on retention of local information. Our results further the understanding of embeddings, their relation to model performance, and shed light on possible future training objectives and optimizing conditional embeddings.
\end{abstract}



\section{Introduction}

Embedding language models, or embedding models, are language models optimized to produce numerical representations of text for a variety of tasks, such as clustering, retrieval, and classification \citep{xie-2016-embeddings-for-clustering, huang-2020-embedding-retrieval, wang-2016-embedding-classification}. Because embeddings are a core component of many current NLP applications, evaluating them---along with studying their structure and optimization methods---is essential. To date, evaluation of embedding models has primarily focused on downstream performance: models are assessed by embedding large datasets, which is both time-consuming and requires large amounts of compute. The application-focused perspective has also left comparatively less attention for understanding \textit{why} one embedding model outperforms another. This gap in understanding which properties are beneficial or detrimental limits our ability to design effective training objectives, select models that generalize beyond test sets, analyze failure cases, and develop new optimization approaches, such as condition-aware embeddings.

We study what makes an embedding model effective by analyzing structure retention. We hypothesize that a model’s suitability for a task is reflected in how it organizes its embedding space. Ideally, task-relevant information is presented systematically, or figuratively on a \textit{silver platter}: for example, translation should benefit from placing parallel sentences in different languages in analogous configurations, in which case moving between languages would correspond to a relatively simple mapping in the space. We apply the same principle to a broader set of embedding tasks.
Rather than comparing embeddings \textit{across} models, we analyze each model’s embedding space in isolation by exploiting the paired structure of common embedding evaluation tasks (e.g., question-answer pairs). This allows us to study what drives task success, rather than only reporting an aggregate evaluation score. Specifically, we investigate the degree to which query and target sides of common embedding benchmark datasets align with each other using $k\NN$-overlap and a linear method based on independent component analysis (ICA) as probes. We uncover strong relationships between embedding space geometry and benchmark performance, and our results provide evidence that these properties can help explain performance differences across tasks. Our contributions are: 

\textbf{Finding 1} We show that high $k$-Nearest Neighbors overlap between queries and targets strongly correlates (even up to 0.97) with performance on the given task. This suggests that strong embedding models organize paired texts in a mutually consistent way in their embedding spaces.

\textbf{Finding 2} We show that the magnitude difference in ICA components (\quot{\textit{peak}}) similarly strongly correlates with performance, and it can be used to analyze the linearity of task information.

\textbf{Finding 3} We tie in these measures by showing the differences in structure they are able to uncover and validate our results using the interpretable features of ICA.

\section{Background and Motivation}

Language model representational spaces, such as embeddings, outputs before next token prediction, or internal states, have been extensively studied in prior work. 
One recurring theme is \textit{linearity}: the idea that concepts can be represented as linear combinations of others. This observation traces back to early word embeddings \citep{mikolov-etal-2013-linguistic} and has since been examined in transformer models \citep{nanda-etal-2023-emergent-linear-representations, gurnee2024LMsRepresentTime&Space, Park2024LinearRepresentations}, with evidence that high-level concepts are often encoded linearly. Consistent with this, embeddings have been successfully analyzed with linear methods, particularly Independent Component Analysis (ICA) \citep{hyvarinen-1999-robustness} in, e.g., \citet{yamagiwa-etal-2023-discovering, li-etal-2024-exploring-intra, musil-marecek-2024-exploring}. Another emerging approach for analyzing representations is the superposition viewpoint \citep{elhage2022toymodelssuperposition}, which argues that concepts can be encoded more efficiently as nearly orthogonal rather than fully orthogonal representations; notable examples include \citet{cunningham2023sparseautoencodershighlyinterpretable, gurnee2023findingNeuronsInHaystackWithProbing, chen-etal-2025-knowledge}. Finally, \citet{park2025hierarchicalgeometry} report that high-level concepts (e.g., language, gender, or animacy) exhibit linearity, whereas lower-level concepts are encoded in a more complex structure.

Several studies show that different embedding models share substantial geometric structure and that their spaces can be aligned with relatively simple mappings. \citet{ren2023roadsleadromeexploring} argue that embedding spaces of differently initialized models are isomorphic up to some tolerance. \citet{yoon-arik-2025-embedding-converter} propose the Embedding Converter, a framework for mapping one embedding space to another. \citet{lin2019situatingsentenceembeddersnearestneighboroverlap} compare models via a nearest neighbor overlap score to quantify shared local structure. In addition, \citet{wang-etal-2022-just-rank} connect embedding space structure to evaluation scores by analyzing exclusively semantic text similarity (STS) tasks and cosine similarity.

\begin{figure}[b]%
    \centering
    \subfloat[\centering Intfloat/multilingual-e5-large-instruct]{{\includegraphics[width=6cm]{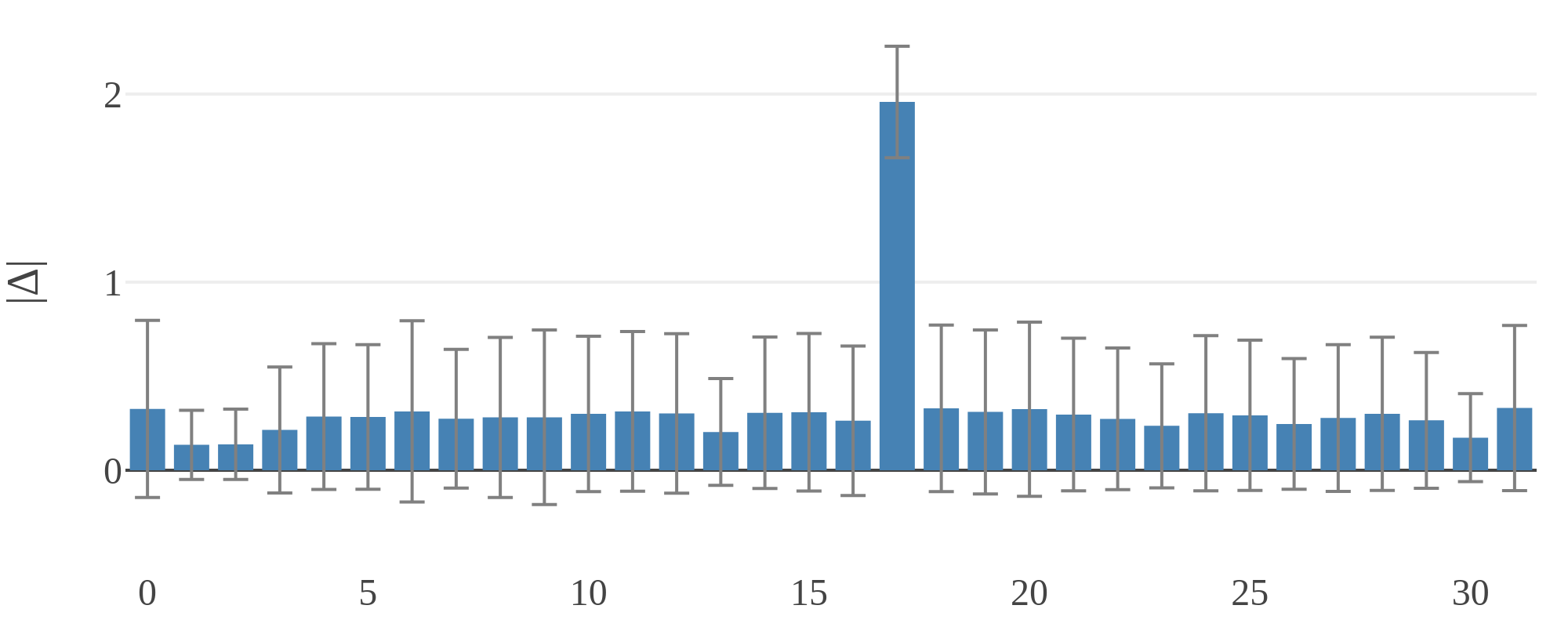} }}%
    \qquad
    \subfloat[\centering  Google/embedding-gemma-300m]{{\includegraphics[width=6cm]{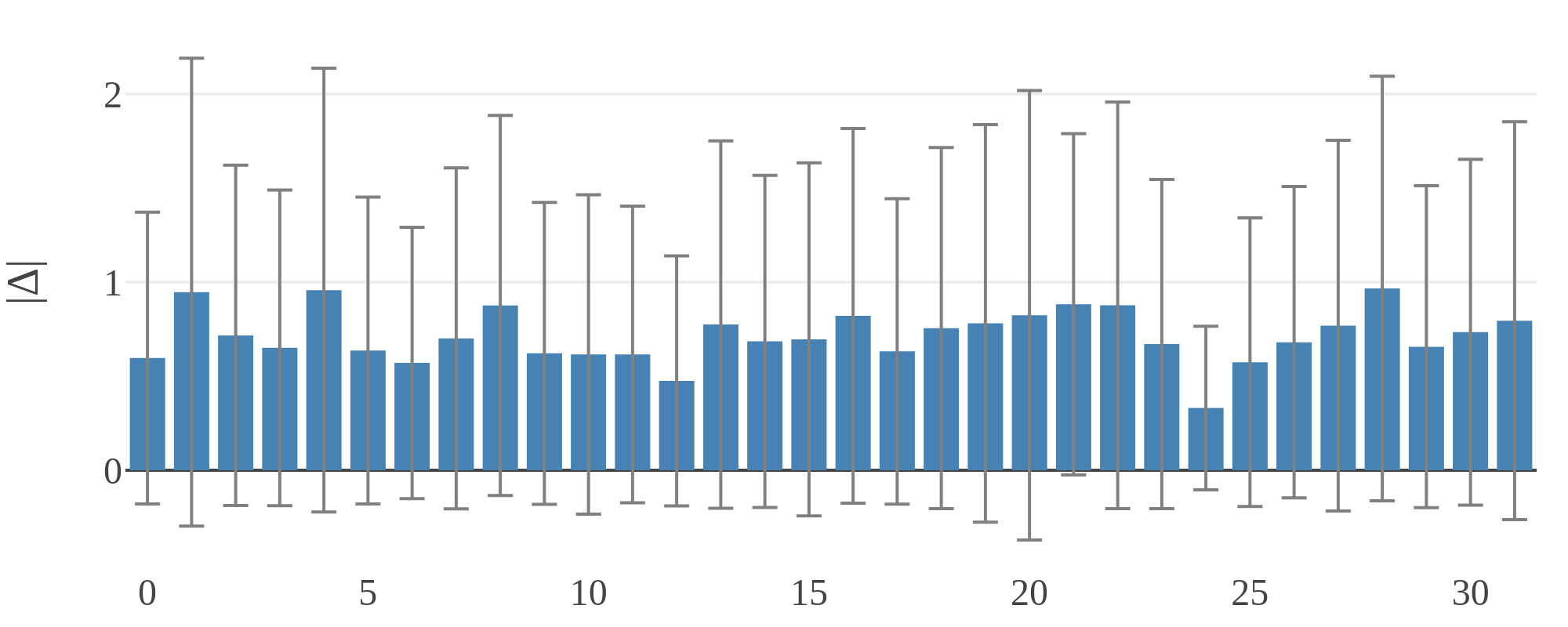} }}%
    \caption{Average absolute difference ($\vert \Delta \vert$) over English-French translation pairs on ICA (dim=32) transformed embeddings per output dimension, with standard deviations as error bars. Multilingual-e5-large instruct, which receives high scores for English-French task, shows a characteristic \quot{peak}, while embedding-gemma-300 does not, and correspondingly receives a lower score for the same task.}%
    \label{fig:peak-example}%
\end{figure}

Our motivation is as follows: if modern embedding models exhibit structures described above, then a simple linear method such as ICA should be able to isolate a dominant concept-level signal (e.g., language) in their embedding spaces. To illustrate the intuition, consider a dataset of $N$ translation pairs $D = \bigl[(t_1,t'_1), (t_2,t'_2),\dots, (t_N,t'_N)\bigr]$ from English texts $t_i$ (queries) to French $t'_i$ (targets). Let $f(t) \in \mathbb R^d$ denote the embedding of $t$. We fit ICA on the embeddings of both sides of the pairs,
\vspace{-4pt}
\begin{align*}
    \text{ICA} \overset{\text{fit}}{\leftarrow} \big\{ f(t_1), \dots, f(t_N), f(t'_1), \dots, f(t'_N)  \bigr\}.
\end{align*}
If language is, as assumed, the primary systematic difference between queries and targets and is represented approximately linearly, ICA should allocate a dedicated component to this signal. Consequently, when we transform both embeddings and take the element-wise difference for each pair, 
$
\text{ICA}\bigl(f(t_i)\bigr) - \text{ICA}\bigl(f(t'_i)\bigr),
$
most of the magnitude of this difference should concentrate in a small number of ICA dimensions. When we plot the mean absolute difference across query and target pairs for each ICA dimension, this concentration should appear as a pronounced \quot{\textit{peak}}. Figure \ref{fig:peak-example} shows this effect for two models, multilingual-e5-large-instruct and embedding-gemma-300m. A peak is clearly observable for multilingual-e5-large-instruct, indicating that ICA isolates a language-specific direction in its output space, and the model coincidentally receives a high score on an English-French task, whereas embedding-gemma-300m does not produce such a peak and receives a lower score. We show that the \textbf{prominence of the peak is descriptive of the performance of the model}.

However, not all concepts are represented linearly in representational spaces (e.g., \citet{park2025hierarchicalgeometry}). Therefore, we also consider a more general notion of structural similarity between queries and targets. Rather than explicitly learning a mapping from $f(t_i)$ to $f(t'_i)$, we probe it directly using the query–target structure of embedding evaluation datasets. Concretely, we compute the \textit{neighborhood retention rate}: the fraction of $k$-nearest neighbors shared between $f(t_i)$ and $f(t'_i)$, and find strong evidence that \textbf{models with higher neighborhood retention also achieve higher evaluation scores}.

In this paper, we widen the scope from the exemplified translation task and target five diverse tasks from four different task categories, showing that the above-described relations between embedding space properties and evaluation scores hold for 25 contemporary embedding models: across these settings, we establish that the two described structural measures correlate strongly with benchmark scores. We further analyze how these measures relate to each other, use ICA to characterize task-specific linear structure, and leverage the interpretability of ICA components to validate the findings.

\section{Methods}

\subsection{Datasets}

The datasets used in this study are ARCChallenge \citep{clark2018ARCChallenge}, WebFAQ \citep{dinzinger2025webfaq}, Tatoeba \citep{tatoeba}, RTE3 \citep{giampiccolo-etal-2007-RTE3}, and SummEval \citep{fabbri2020summeval}. A summary of these datasets is presented in Table \ref{tab:datasets}, and the languages included in multilingual tasks are shown in the result tables (e.g., Table \ref{tab:neighbor_results}) as ISO language codes. The datasets cover four task types in the MTEB benchmark, and were selected because they contain well-defined paired text instances. For example, ARCChallenge contains question–answer pairs, SummEval document–summary pairs; we refer to these partitions as queries and targets, respectively.
We use the test/development sets and give further dataset construction details in Appendix \ref{app:data-sampling}, such as the selection of contradicting passages for RTE3.

\subsection{Embedding Models}

Embedding models used in this study are listed in Table \ref{tab:models}.
We select models that vary along several dimensions to ensure the soundness of our experiments: (1) scores on the MTEB multilingual leaderboard (spanning a wide range); (2) model developer (covering a diverse set, with a few models chosen per family); (3) model accessibility (primarily open-weight models, with a few prominent closed-weight models); and (4) language scope (including both English-only and multilingual models). This selection helps ensure that observed differences reflect actual model capability variation rather than only confounding factors.
Additionally, most contemporary embedding models are intended to be used with a prompt, a short instruction added to the query to help steer the embedding towards the desired output. We therefore run our main experiments both with and without prompts. 
The prompts, the prompt template, and additional model information are presented in Appendix \ref{app:prompting}. We use the \texttt{sentence-transformers} \citep{reimers-2019-sentence-transformers} library.

\begin{table}[b]
    \caption{Datasets used in our study. $N$ denotes the number of paired instances.}
    \label{tab:datasets}
    \centering
    \begin{tabular}{lllll}
        \toprule
        Dataset name & MTEB category & Pair difference & Language(s) & $N$\\ \midrule
        ARCChallenge 
        & Retrieval & Question--Answer & English & 937\\
        WebFAQ 
        & Retrieval & Question--Answer & 13 languages & 5000\\
        Tatoeba 
        & Bitext Mining & English--Non-English & 10 languages & 5000\\
        RTE3 
        & Pair Classification & Premise--Contradiction & 4 languages & 90 \\
        SummEval 
        & Summarisation & Document--Summary  & English & 80 \\  \bottomrule
    \end{tabular}
\end{table}

\subsection{MTEB Evaluation}

We use the evaluation scores from the official \rhref{MTEB repository}{https://github.com/embeddings-benchmark/mteb} (accessed 12th Feb. 2026). 
However, this repository only contains language-aggregated score for some tasks, in which case we evaluate the tasks ourselves with the \texttt{mteb}-library \citep{muennighoff-2023-mteb, enevoldsen2025mmteb} with default settings.
Unfortunately, the scores for WebFAQ are not available for closed-source models, nor is it feasible to evaluate this task on them due to the large number of examples (>100k per language). Similarly, evaluating WebFAQ on the largest models using \texttt{mteb}-library would require computational resources beyond our reach, so for this specific task, we report the results for a truncated set of models, with the omitted models shown in Table \ref{tab:models}.

\begin{table}[]
    \centering
    \caption{Embedding models used in this study, number of parameters (Par.) in units of billion, and average score (Sc.) on the MTEB multilingual leaderboard. Models are categorized by provider or base model from which the model is finetuned. Models marked with a star (*) are not included in the WebFAQ evaluation, and models with no parameter counts are closed-weight models.}
   \begin{tabular}{@{}lccllcc@{}}
\toprule
Model & Par. (B) & Sc. $\uparrow$ &  & Model   & Par. (B) & Sc. $\uparrow$ \\ \cmidrule(r){1-3} \cmidrule(l){5-7} 
\rhref{bge-m3}{https://huggingface.co/BAAI/bge-m3}                                                     & 0.568    & 59.56            &  & \rhref{granite-107m-multilingual}{https://huggingface.co/ibm-granite/granite-embedding-107m-multilingual} & 0.107   & 51.81            \\
\rhref{bge-base-en-v1.5}{https://huggingface.co/BAAI/bge-base-en-v1.5}                                 & 0.033    & 43.76            &  & \rhref{granite-125m-english}{https://huggingface.co/ibm-granite/granite-embedding-125m-english}           & 0.125   & 44.04            \\ \cmidrule(r){1-3} \cmidrule(l){5-7} 
\rhref{jina-b-en-v1}{https://huggingface.co/jinaai/jina-embedding-b-en-v1}                   & 0.11     & 40.91            &  & \rhref{gemini-embedding-001}{https://ai.google.dev/gemini-api/docs/embeddings}*                                     & -       & 68.37            \\
\rhref{jina-v2-small-en}{https://huggingface.co/jinaai/jina-embeddings-v2-small-en}         & 0.033    & 35.22            &  & \rhref{EmbeddingGemma}{https://ai.google.dev/gemma/docs/embeddinggemma}                                             & 0.308   & 61.15            \\ \cmidrule(r){1-3}
\rhref{multilingual-e5-large-inst.}{https://huggingface.co/intfloat/multilingual-e5-large-instruct} & 0.56     & 63.22            &  & \rhref{LaBSE}{https://www.kaggle.com/models/google/labse/tensorFlow2/labse/1?tfhub-redirect=true}                   & 0.471   & 52.07            \\ \cmidrule(l){5-7} 
\rhref{e5-mistral-7b-instruct}{https://huggingface.co/intfloat/e5-mistral-7b-instruct}*                & 7.111    & 60.25            &  & \rhref{Qwen3-Embedding-8B}{https://huggingface.co/Qwen/Qwen3-Embedding-8B}*                                         & 7.567   & 70.58            \\
\rhref{multilingual-e5-large}{https://huggingface.co/intfloat/multilingual-e5-large}                   & 0.118    & 58.62            &  & \rhref{Qwen3-Embedding-0.6B}{https://huggingface.co/Qwen/Qwen3-Embedding-0.6B}*                                     & 0.596   & 64.34            \\
\rhref{e5-small}{https://huggingface.co/intfloat/e5-small}                                             & 0.033    & 43.46            &  & \rhref{gte-multilingual-base}{https://huggingface.co/Alibaba-NLP/gte-multilingual-base}                             & 0.305   & 58.34            \\ \cmidrule(r){1-3}
\rhref{text-embedding-3-large}{https://openai.com/index/new-embedding-models-and-api-updates/}*        & -        & 58.96            &  & \rhref{KaLM-multilingual-mini}{https://huggingface.co/HIT-TMG/KaLM-embedding-multilingual-mini-v1}     & 0.494   & 57.05            \\
\rhref{text-embedding-3-small}{https://openai.com/index/new-embedding-models-and-api-updates/} *       & -        & 54.01            &  & \rhref{gte-small}{https://huggingface.co/thenlper/gte-small}                                                        & 0.033   & 44.1             \\ \cmidrule(r){1-3} \cmidrule(l){5-7} 
\rhref{potion-multilingual-128M}{https://huggingface.co/minishlab/potion-multilingual-128M}            & 0.128    & 47.23            &  & \rhref{llama-embed-nemotron-8b}{https://huggingface.co/nvidia/llama-embed-nemotron-8b}*                             & 7.505   & 69.46            \\ \cmidrule(l){5-7} 
\rhref{potion-base-8M}{https://huggingface.co/minishlab/potion-base-8M}                                & 0.008    & 38.56            &  & \rhref{snowflake-arctic-l-v2.0}{https://huggingface.co/Snowflake/snowflake-arctic-embed-l-v2.0}               & 0.568   & 57.03            \\
   &          &                  &  & \rhref{snowflake-arctic-l}{https://huggingface.co/Snowflake/snowflake-arctic-embed-l}                         & 0.335   & 43.33            \\ \bottomrule
\end{tabular}
    \label{tab:models}
\end{table}

\subsection{Neighborhood Retention}

We evaluate the neighborhood retention between the query and target $(t, t')$ as
\vspace{-2pt}
\[\text{Ret}_k = \sum_{(t,t')\in D} \frac{\text{Jacc}\left(k\NN(f(t)) , k\NN(f(t'))\right)}{k}\]
where $k\NN(f(t))$ is the set of $k$ nearest neighbors for the embedding of text $t$ and $\text{Jacc}(\cdot, \cdot)$ is the Jaccard-overlap. This quantity is inspired by the local training loss of \citet{yoon-arik-2025-embedding-converter}, which they use to preserve the local structure in their Embedding-Converter, and the N2O metric used by \citet{lin2019situatingsentenceembeddersnearestneighboroverlap}. We roughly optimize $k$ with a grid search covering $k = \{5,10, 20, 50\} \cup \{ 0.01N, 0.02N, 0.05N, 0.1N , 0.2N\}$ where $N$ is the number of paired instance in the dataset $D$, accounting for diversity in dataset sizes. We report the results for $k = 10$, which was found to be in the top 3 values for all datasets. It is to be noted that while the value of $k$ has an impact on the results, \textbf{all tested values of $k$ support our conclusions} (see Appendix \ref{app:bad_k}).

\subsection{Independent Component Analysis and Peak Prominence}

Independent component analysis \citep{hyvarinen-1999-robustness} is a linear blind source separation algorithm. 
A more detailed explanation of ICA is presented in Appendix \ref{app:ICA}. We use ICA in this study for three reasons: we aim to evaluate linearity, it offers straightforward interpretability, and fitting the required number of ICA models for our experiments is substantially less costly than training the same number of other interpretable models, such as Sparse Autoencoders (SAEs). We use the FastICA algorithm from the \texttt{sklearn}-library \citep{pedregosa-2011-scikitlearn} with default parameters, except for the maximum number of iterations, which we set to 10{,}000 to ensure convergence. 

In our experiments, we fit the ICA models, subtract paired instances, and analyze the prominence of the characteristic peak with the Gini coefficient: a statistic measuring the inequality of a distribution, 0 indicating a uniform distribution, and 1 a totally unequal distribution. Using the Gini coefficient was motivated by the level of agreement between Gini values and visual inspection of peaks, and its stability against other suitable metrics (e.g., peak-to-mean ratio), as evaluated in Appendix \ref{sec:stability}.
While ICA is developed for settings where the number of sources is known upfront, it has been successfully applied to embeddings with a smaller number of assumed sources $d_{\text{ICA}}$ \citep{huertas-garcia-2023-dimensionality-reduction}. In preliminary experiments, we observed that large values of $d_{\text{ICA}}$ produced no peaks or failed to converge, while smaller values led to a clear signal, which we attribute to the homogeneity and small size of our datasets. We therefore set $d_{\text{ICA}} \ll d_{\text{emb}}$ and compare $\{32, 64, 128\}$ and PCA-estimated number of dimensions accounting 50\% and 90\% of explained variance. The clearest peaks occur at 32 and 64 dimensions, so we report all results for ICA$^{32}$ (and ICA$^{64}$ in Appendix \ref{app:ICA64}).

\section{Experiments}

We divide our experiments into two main experiments examining the relationship between embedding space structure and benchmark scores, and supporting experiments that provide interpretability and validate the main results. The computational cost of our experiments is reported in Appendix \ref{app:compute}.

In our first main experiment, we compute the neighborhood retention score $\text{Ret}_k$ for all model–dataset combinations and correlate it with MTEB performance. This tests whether organizing the embedding space locally similarly on the query and target sides is beneficial for solving the task.
In the second main experiment, we fit an ICA model on the embeddings, transform both sides of each pair, and compute the element-wise mean absolute difference between paired transformed embeddings. This produces a distribution over ICA dimensions with a varying \quot{peak} (see Figure \ref{fig:peak-example}). We quantify peak prominence using the Gini coefficient and correlate it with MTEB performance. This experiment tests whether the query–target difference concentrates in a small number of independent directions, consistent with task-relevant information being encoded in an approximately linear form, and whether this property is beneficial for model performance. For both main experiments, we report Spearman correlation coefficients, since in each case we compare two rankings.

To support the main findings, we further examine how neighborhood retention relates to ICA by interpreting the ICA components in terms of local (pair-specific) and global (dataset-level) signals shared between queries and targets. Additionally, in the Appendix, we validate our approach by (1) examining the stability of the fitted ICA models across runs with different random initializations (Appendix \ref{sec:stability}); and (2) extracting the most salient words for each ICA dimension to verify that the observed peak corresponds to interpretable linguistic content rather than a spurious effect (Appendix \ref{sec:keywords+matrix}). All supporting experiments are described in more detail in their respective sections.

\begin{table}[t]
    \centering
    \caption{Spearman correlation ($\uparrow$) for neighborhood retention with MTEB scores, with bolded values for strong relationship ($\geq 0.7$). Star ($*$) indicates significance at level $p<0.05$, dagger ($\dagger$) at $p<0.01$, and double dagger ($\ddagger$) at $p<0.001$.}
    \label{tab:neighbor_results}
\begin{tabular}{lccllcc}
\toprule
Dataset         & Unprompted                 & Prompted                   &  & Dataset    & Unprompted                 & Prompted                   \\ \cline{1-3} \cline{5-7} 
ARCChallenge  & \textbf{0.890}$^{\ddagger}$ & \textbf{0.945}$^{\ddagger}$  & & SummEval  & 0.643$^{\ddagger}$ & 0.588$^{\dagger}$ \\ \cline{1-3} \cline{5-7}
RTE3:eng  & -0.230 & -0.058  & & WebFAQ:bul  & \textbf{0.926}$^{\ddagger}$ & \textbf{0.866}$^{\ddagger}$ \\
RTE3:fra  & 0.149 & 0.268  & & WebFAQ:deu  & \textbf{0.889}$^{\ddagger}$ & \textbf{0.856}$^{\ddagger}$ \\
RTE3:ita  & 0.389 & 0.488$^{*}$  & & WebFAQ:ell  & \textbf{0.946}$^{\ddagger}$ & \textbf{0.930}$^{\ddagger}$ \\ 
RTE3:deu  & 0.025 & 0.162  & & WebFAQ:eng  & 0.620$^{\dagger}$ & 0.544$^{*}$ \\ \cline{1-3}
Tatoeba:deu-eng  & \textbf{0.888}$^{\ddagger}$ & \textbf{0.904}$^{\ddagger}$  & & WebFAQ:fas  & \textbf{0.934}$^{\ddagger}$ & \textbf{0.893}$^{\ddagger}$ \\
Tatoeba:bul-eng  & \textbf{0.962}$^{\ddagger}$ & \textbf{0.928}$^{\ddagger}$  & & WebFAQ:fin  & \textbf{0.953}$^{\ddagger}$ & \textbf{0.913}$^{\ddagger}$ \\
Tatoeba:ukr-eng  & \textbf{0.939}$^{\ddagger}$ & \textbf{0.918}$^{\ddagger}$  & & WebFAQ:fra  & \textbf{0.878}$^{\ddagger}$ & \textbf{0.843}$^{\ddagger}$ \\
Tatoeba:ita-eng  & \textbf{0.917}$^{\ddagger}$ & \textbf{0.884}$^{\ddagger}$  & & WebFAQ:hin  & \textbf{0.909}$^{\ddagger}$ & \textbf{0.891}$^{\ddagger}$ \\
Tatoeba:fra-eng  & \textbf{0.925}$^{\ddagger}$ & \textbf{0.895}$^{\ddagger}$  & & WebFAQ:ita  & \textbf{0.907}$^{\ddagger}$ & \textbf{0.816}$^{\ddagger}$ \\
Tatoeba:vie-eng  & \textbf{0.945}$^{\ddagger}$ & \textbf{0.916}$^{\ddagger}$  & & WebFAQ:jpn  & \textbf{0.864}$^{\ddagger}$ & \textbf{0.827}$^{\ddagger}$ \\
Tatoeba:ell-eng  & \textbf{0.968}$^{\ddagger}$ & \textbf{0.922}$^{\ddagger}$  & & WebFAQ:ukr  & \textbf{0.942}$^{\ddagger}$ & \textbf{0.909}$^{\ddagger}$ \\
Tatoeba:jpn-eng  & \textbf{0.968}$^{\ddagger}$ & \textbf{0.939}$^{\ddagger}$  & & WebFAQ:vie  & \textbf{0.936}$^{\ddagger}$ & \textbf{0.903}$^{\ddagger}$ \\
Tatoeba:hin-eng  & \textbf{0.931}$^{\ddagger}$ & \textbf{0.876}$^{\ddagger}$  & & WebFAQ:zho  & \textbf{0.891}$^{\ddagger}$ & \textbf{0.841}$^{\ddagger}$ \\
Tatoeba:fin-eng  & \textbf{0.968}$^{\ddagger}$ & \textbf{0.959}$^{\ddagger}$  & & \\ \bottomrule
\end{tabular}
\end{table}

\section{Results}
\label{sec:results}

\begin{figure}[t]
    \centering
    \scriptsize
        \includegraphics[width=0.9\linewidth]{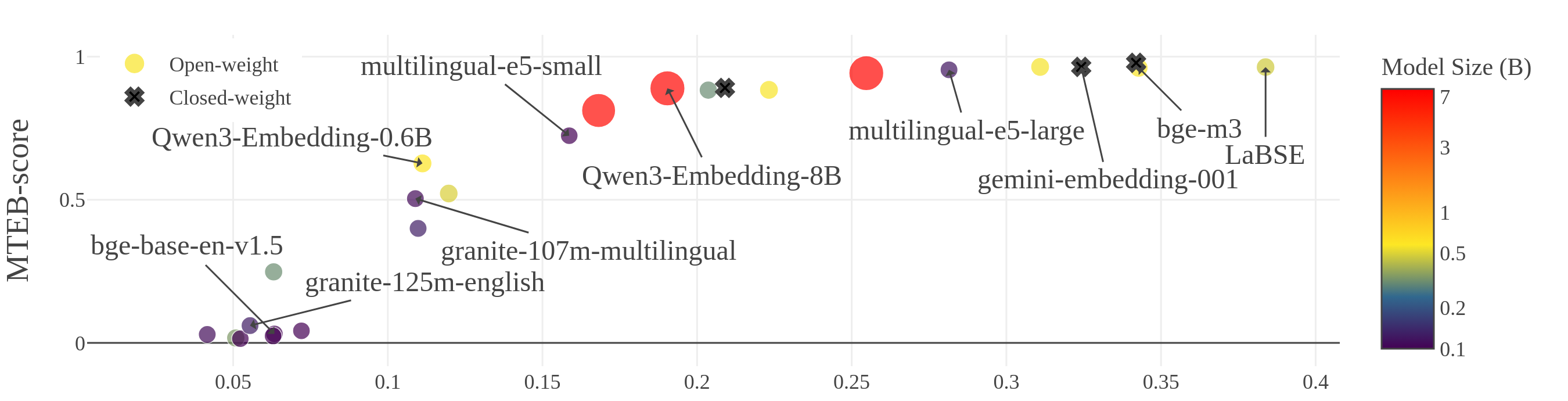}
    \caption{Visualization of the relationship between $\text{Ret}_k$ (horizontal) and the MTEB-score for Tatoeba:fin-eng task, which shows one of the strongest correlations in Tatoeba (\textbf{0.968}$^\ddagger$). Annotated model names show that multilingual models outperform their English-only counterparts, and that model size is not the only contributing factor. Other examples in Appendix \ref{app:complementary_figures}.}
    \label{fig:annotated_knn_example}
\end{figure}

\textbf{Neighborhood Retention vs. Performance} \ \ The results for neighborhood retention against MTEB-score are presented in Table \ref{tab:neighbor_results}. Figure \ref{fig:annotated_knn_example} presents an example of the relationship between retention and score. Overall, neighborhood retention correlates strongly with MTEB performance on three of the five datasets and moderately on one, with significance of $p<0.001$ in almost all cases where a strong relationship is found. This suggests that embedding models that preserve local neighborhood structure between paired inputs for a given task tend to perform well on that task, whereas models with less consistent structure tend to perform worse. For Tatoeba, the relationship is close to perfect ($\geq 0.9$) with only a few exceptions. ARCChallenge and WebFAQ, both question-answering datasets, show similarly strong correlations (outside WebFAQ:eng, see Appendix Figure \ref{fig:knn:WebFAQ:eng}). SummEval shows a modest but significant correlation, whereas RTE3 shows little to no correlation. The effect of prompting varies by dataset: it is most beneficial for ARCChallenge, likely because our ARCChallenge prompt matches the one used in MTEB and therefore best reflects the intended evaluation setup. In contrast, prompting slightly reduces the correlation for other datasets.

\begin{table}[]
    \centering
    \caption{Spearman correlation ($\uparrow$) for Gini-coefficient calculated from ICA$^{32}$ with MTEB scores, with bolded values for strong relationship ($\geq 0.7$). Star ($*$) indicates significance at level $p<0.05$, dagger ($\dagger$) at level $p<0.01$, and double dagger ($\ddagger$) at level $p<0.001$.}
    \label{tab:gini_ica32}
   \begin{tabular}{lccllcc}
\toprule
Dataset         & Unprompted                  & Prompted                    &  & Dataset    & Unprompted         & Prompted           \\ \cline{1-3} \cline{5-7} 
ARCChallenge  & \textbf{0.810}$^{\ddagger}$ & \textbf{0.840}$^{\ddagger}$  & & SummEval  & 0.178 & 0.428$^{*}$ \\ \cline{1-3} \cline{5-7} 
RTE3:eng  & 0.414$^{*}$ & 0.318  & & WebFAQ:bul  & 0.286 & 0.278 \\
RTE3:fra  & \textbf{0.706}$^{\ddagger}$ & 0.542$^{\dagger}$  & & WebFAQ:deu  & \textbf{0.756}$^{\ddagger}$ & 0.680$^{\dagger}$ \\
RTE3:ita  & 0.652$^{\ddagger}$ & 0.541$^{\dagger}$  & & WebFAQ:ell  & -0.106 & 0.284 \\
RTE3:deu  & 0.525$^{\dagger}$ & 0.555$^{\dagger}$  & & WebFAQ:eng  & 0.255 & 0.542$^{*}$ \\ \cline{1-3}
Tatoeba:deu-eng  & \textbf{0.818}$^{\ddagger}$ & \textbf{0.933}$^{\ddagger}$  & & WebFAQ:fas  & -0.158 & -0.133 \\
Tatoeba:bul-eng  & \textbf{0.861}$^{\ddagger}$ & \textbf{0.928}$^{\ddagger}$  & & WebFAQ:fin  & 0.680$^{\dagger}$ & \textbf{0.736}$^{\ddagger}$ \\
Tatoeba:ukr-eng  & \textbf{0.906}$^{\ddagger}$ & \textbf{0.918}$^{\ddagger}$  & & WebFAQ:fra  & \textbf{0.721}$^{\ddagger}$ & 0.622$^{\dagger}$ \\
Tatoeba:ita-eng  & \textbf{0.942}$^{\ddagger}$ & \textbf{0.909}$^{\ddagger}$  & & WebFAQ:hin  & 0.152 & 0.183 \\
Tatoeba:fra-eng  & \textbf{0.796}$^{\ddagger}$ & \textbf{0.859}$^{\ddagger}$  & & WebFAQ:ita  & 0.626$^{\dagger}$ & 0.564$^{*}$ \\
Tatoeba:vie-eng  & \textbf{0.829}$^{\ddagger}$ & \textbf{0.881}$^{\ddagger}$  & & WebFAQ:jpn  & 0.643$^{\dagger}$ & 0.614$^{\dagger}$ \\
Tatoeba:ell-eng  & \textbf{0.904}$^{\ddagger}$ & \textbf{0.868}$^{\ddagger}$  & & WebFAQ:ukr  & 0.360 & 0.354 \\
Tatoeba:jpn-eng  & \textbf{0.853}$^{\ddagger}$ & \textbf{0.866}$^{\ddagger}$  & & WebFAQ:vie  & 0.558$^{*}$ & 0.657$^{\dagger}$ \\
Tatoeba:hin-eng  & \textbf{0.881}$^{\ddagger}$ & \textbf{0.893}$^{\ddagger}$  & & WebFAQ:zho  & 0.257 & 0.422 \\
Tatoeba:fin-eng  & \textbf{0.945}$^{\ddagger}$ & \textbf{0.928}$^{\ddagger}$  & & \\ \bottomrule       
\end{tabular}
\end{table}

\textbf{Peak Prominence vs. Performance} \ \ The results for Gini-coefficient of ICA$^{32}$ (peak prominence) versus MTEB-score are presented in Table \ref{tab:gini_ica32}. Tatoeba shows correlations comparable to those observed in the neighborhood retention experiment, indicating that bitext mining benefits from preserving the relative neighborhood organization between the two sides of each pair, and that a more linear encoding of translation-pair differences is characteristic of high-performing models. ARCChallenge also shows this connection to a lesser degree, and again benefits from prompting. The main differences between neighborhood retention and peak prominence emerge for RTE3 and WebFAQ, but in opposite directions: Gini yields a stronger relationship for RTE3, whereas correlations for WebFAQ are weaker than in the neighborhood retention results, even when they remain significant and moderately positive in half of the studied languages. The results for ICA$^{64}$ are presented in Appendix \ref{app:ICA64}, and they support the same conclusions outside prompting.

In summary, although the tasks differ in how strongly structure relates to performance, our results show that simple measures computed from paired embeddings can correlate strongly with benchmark scores. Comparing neighborhood retention and ICA, we find that on the Tatoeba dataset, both measures achieve very high correlations, consistent with prior work suggesting that language is encoded more linearly than many lower-level concepts. We also show that it is possible to probe this quality using ICA. For ARCChallenge, our most faithful prompt evaluation, prompting improves the correlation, leading us to draw the conclusion that adding a prompt modifies the query with a transformation related to local structure retention between queries and targets.
For other datasets, such as WebFAQ, neighborhood retention reliably predicts model ranking, but linear ICA does not capture the relationship consistently across languages. This points to cross-lingual differences in how question–answer structure is encoded in the embedding space. Finally, unlike neighborhood retention, ICA reveals a meaningful relationship for RTE3, which we analyze further in the next section.

\begin{figure}[!ht]%
    \centering
    \scriptsize
    \subfloat[\centering  Tatoeba:eng-ell ( \textcolor{brickred}{$\bullet$} 0.904$^\ddagger$ vs. \textcolor{navyblue}{$\blacklozenge$} 0.218)]
    {
        {\includegraphics[width=0.4\linewidth]{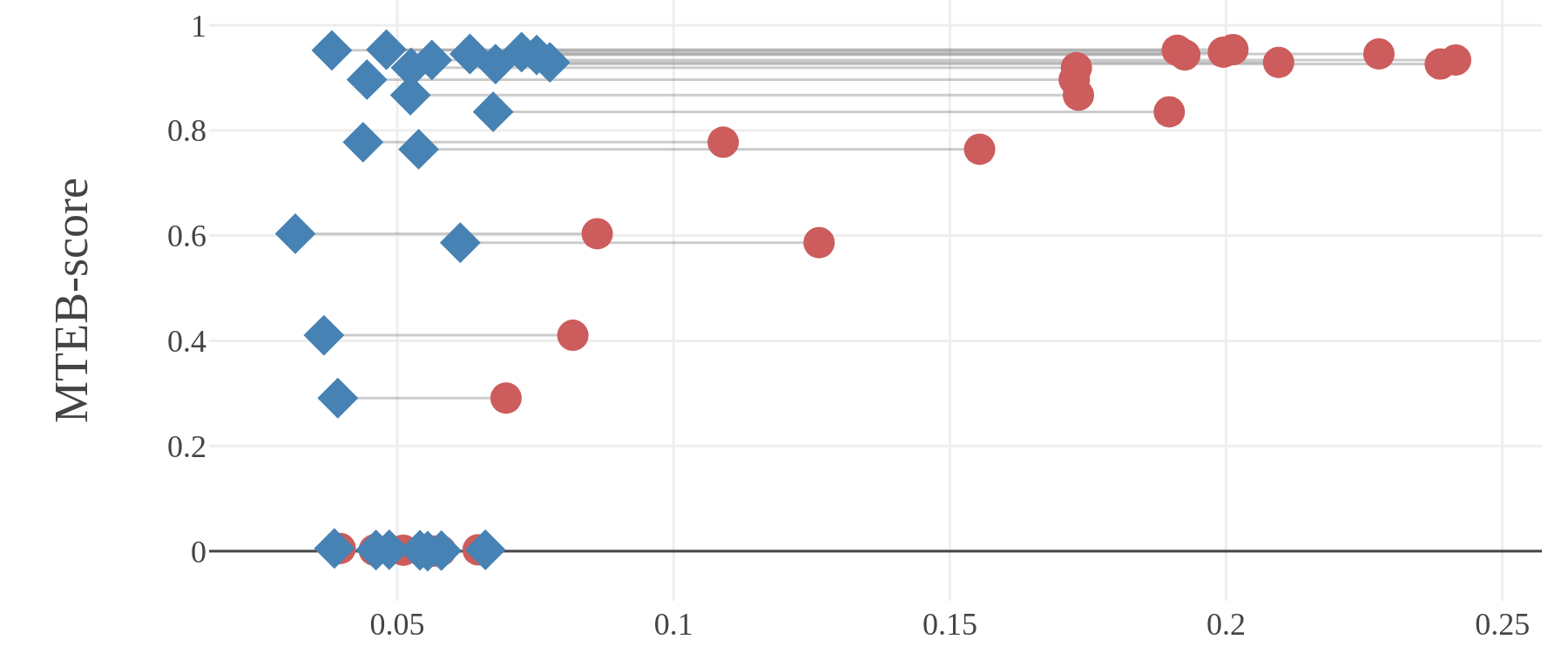} }
    }
    \quad \quad
    \subfloat[\centering RTE3:eng (\textcolor{brickred}{$\bullet$} 0.414$^*$ vs. \textcolor{navyblue}{$\blacklozenge$} 0.436$^*$)]
    {
        {\includegraphics[width=0.41\linewidth]{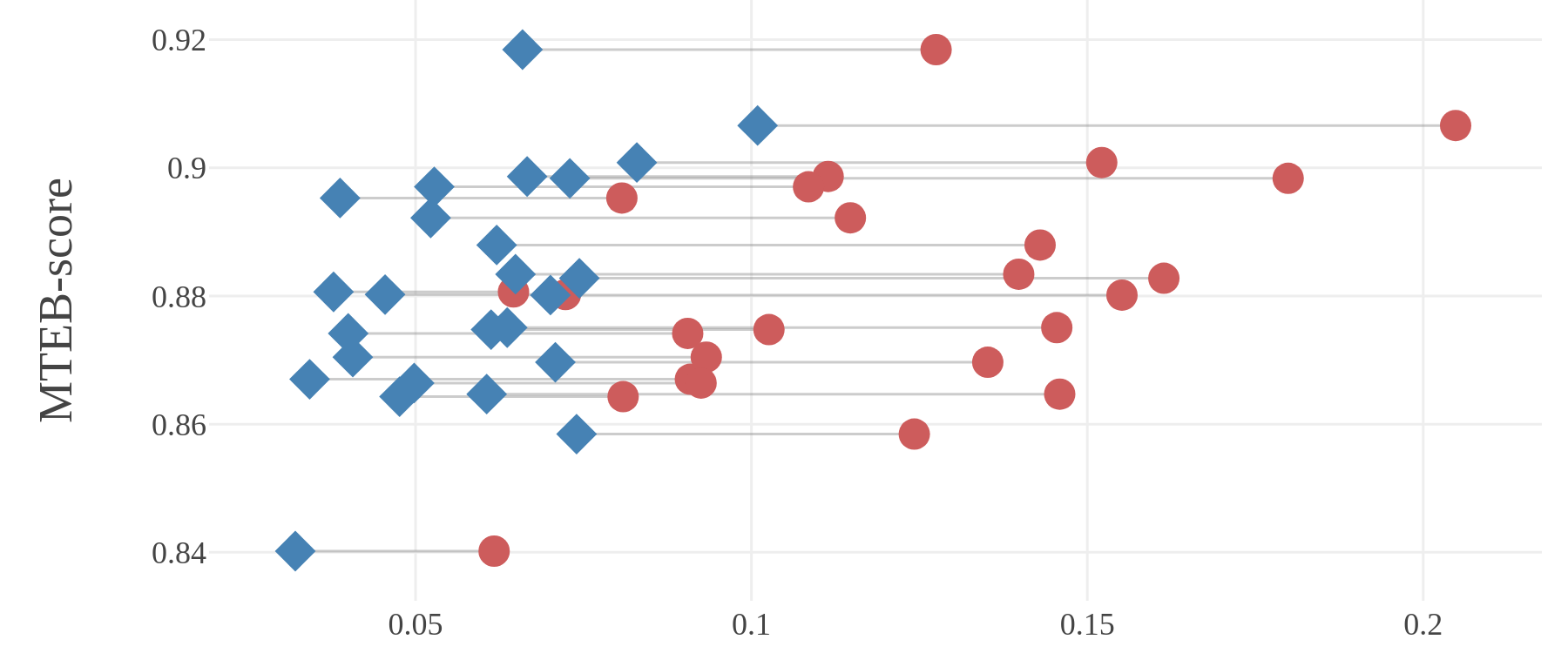} }
    }
    \caption{Comparison between paired (\textcolor{brickred}{$\bullet$}) and shuffled (\textcolor{navyblue}{$\blacklozenge$}) Gini-coefficient (horizontal) for selected datasets, with the associated correlation coefficient and significance. The effect varies: in Tatoeba, the difference grows along MTEB performance and shuffling destroys the relationship, while in RTE3, the difference doesn't depend on the performance and the correlation actually increases.}%
    \label{fig:shuffling}%
\end{figure}

\textbf{Shuffling ICA} \ \ A pertinent question is what, exactly, ICA captures in our experiment. We address this by separating \textit{local} and \textit{global} signals: we randomly shuffle the targets before computing differences, thereby breaking the correspondence between the two texts in each pair. For example, in Tatoeba, shuffling removes the semantic alignment between translations (a local signal) while preserving the overall difference between English and the other language (a global signal). By comparing results from the paired setting (main experiment) to the shuffled setting (this experiment), we can estimate whether ICA primarily captures pair-specific structure (local) or dataset-level differences (global), and how this relates to task performance. In practice, we reuse the ICA models fitted in the main experiment, but shuffle before subtracting the pairs. Figure \ref{fig:shuffling} shows examples for two datasets, Tatoeba:eng-ell and RTE3:eng, which exhibit qualitatively different behavior under shuffling. In both cases, shuffling reduces the prominence of the ICA peak: the Gini coefficient decreases as expected because the shuffled differences align less consistently with any single extracted ICA component. However, the magnitude of this decrease is dataset-dependent, and the extent to which shuffling disrupts the correlation with MTEB scores differs markedly between shuffled Tatoeba and RTE3.

For Tatoeba, shuffling largely destroys the relationship between the Gini-coefficient and MTEB performance (from \textbf{0.904}$^\ddagger$ to 0.218). After shuffling, all models yield similar Gini values, and the highest-performing models are affected the most. This indicates that ICA primarily captures a \textit{local} signal tied to the correct pairing between each query and its translation—precisely the type of structure that also drives neighborhood retention. This agreement between ICA and neighborhood retention is consistent with prior work suggesting that language-related differences are often encoded in relatively linear directions in embedding and LLM representational spaces.
For RTE3, an opposite pattern emerges: the correlation remains essentially unchanged after shuffling (from 0.414$^*$ to 0.436$^*$), and all models are affected in broadly similar ways. This suggests that, for RTE3, ICA mainly captures a \textit{global} signal that does not depend on the specific query–target pairing. This aligns with the results of neighborhood retention, where local structure did not predict benchmark ranking. In the MTEB evaluation of RTE3, paired instances are labeled based on their relative distance (entailing pairs are close, contradicting pairs are far). Because we evaluate only contradicting pairs (a selection ensuring our experiments measure a variety of non-STS qualities), MTEB effectively rewards models that separate the two texts sufficiently in the embedding space; to receive a high MTEB score, pairwise structure is less critical than achieving a large overall separation, which, in turn, ICA can extract.

\begin{figure}[]%
\scriptsize
\small
    \centering
    \subfloat[\centering High peak observed, paired ($\uparrow$) and shuffled ($\downarrow$)]{
    {\includegraphics[width=0.39\linewidth]{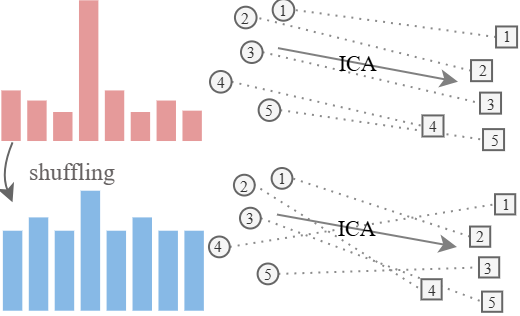}}
    }%
    \qquad \qquad
    \subfloat[\centering Low peak observed, paired ($\uparrow$) and shuffled ($\downarrow$)]{
    
    {\includegraphics[width=0.39\linewidth]{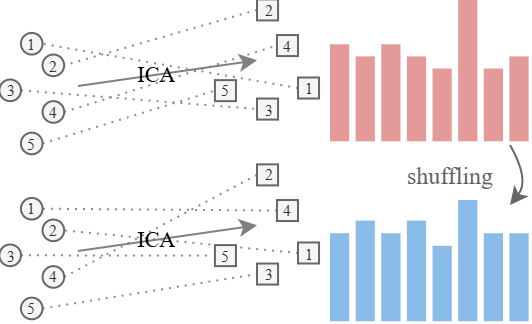}}
    }%
    \caption{Connection between neighbor retention and ICA peaks: When local structure is retained, the individual differences between paired queries (circles) and targets (squares) align with the vector corresponding to the peak extracted by ICA (arrow), and therefore shuffling destroys the peak. When no local structure is retained between queries and targets, the effects of shuffling are minor.}%
    \label{fig:shuffling-vs-neighbors-example}%
\end{figure}

We illustrate the likely mechanism in Figure \ref{fig:shuffling-vs-neighbors-example}. When an embedding space preserves pair-specific local structure between queries and targets, the query–target differences align with a consistent ICA direction, pertaining to a salient peak; shuffling then disrupts this alignment and substantially reduces peak prominence. Conversely, when ICA primarily captures a global dataset-level shift shared across many pairs, the peak is already weaker, and shuffling has little effect. This observation helps reconcile the behaviors of neighborhood retention and the Gini-based ICA measure across tasks. When the query–target relationship is approximately linear, both neighborhood retention and ICA tend to be strong: local neighborhoods are preserved, and the difference concentrates into a small number of ICA directions. When the relationship is strongly non-linear, neighborhood retention may remain high, but ICA captures only part of the structure, leading to weaker or less consistent Gini–performance correlations. Finally, even when neighborhood retention is low (limited local alignment), ICA can still correlate with performance if the benchmark rewards a global separation signal, as in RTE3. The remaining datasets lie between these extremes and exhibit mixed contributions of local and global structure, with selected examples visualized in Appendix Figure \ref{fig:shuffling:ARC+SummEval}.

\textbf{Supporting experiments} \ \ Our experiments rely on using non-deterministic ICA models. We analyze the stability of ICA and the word-embedding explanations of the ICA peak dimensions in Appendices \ref{sec:stability} and \ref{sec:keywords+matrix}, respectively. To summarize, these results show that ICA is able to reliably extract the same underlying representations across different initializations, and that the peak dimensions correspond to words associated with the given tasks. Together, these supporting experiments validate that ICA consistently captures the same underlying signals that our main analysis attributes to it.

\section{Discussion and Related Work}

In this paper, we furthered the understanding of which underlying properties of embeddings lead to high performance. We investigated two methods of probing structural retention, neighborhood retention and ICA. Our results showed that common embedding evaluation datasets display differences in the structural similarities between the query and target side, but that, in general, embedding models that organize their spaces in an analogous geometric fashion exhibit better performance. Taken together, these results indicate that some prevalent tasks reward models whose query--target relationship is well captured by a small number of linear directions. We simultaneously evaluated the degree of linearity the embeddings exhibit in each evaluation dataset. We uncovered connections between ICA and neighborhood retention and showed how linearity and local-global directions in ICA interoperate: the shuffling results show that ICA can reflect either (i) \textit{local} signals tied to correct pairing or (ii) \textit{global} shifts that do not depend on the specific pairing. This helps explain when neighborhood retention and ICA agree (strong local alignment and approximately linear structure) and when they diverge (non-linear local structure or benchmarks that emphasize global separation). Beyond these structural effects, we also observe the effects of prompting (most notably ARCChallenge), highlighting that evaluation choices that alter the query representation can affect not only absolute benchmark scores but also the relationship between geometry and performance. 

Overall, our results provide evidence that embedding space geometry---the consistency of local neighborhoods between paired inputs and the extent to which pairwise differences concentrate into a small number of directions---is closely tied to benchmark performance in several prominent tasks. We found that probes such as neighborhood retention and ICA peakiness can complement downstream evaluation by helping diagnose \textit{how} and \textit{why} models succeed or fail, and offer lightweight signals that may guide future work on training objectives and task- or condition-aware embedding models.


Our work relies on other previous work on analyzing embedding models. Studies using ICA to analyze embeddings include: \citet{yamagiwa-etal-2023-discovering}, \citet{li-etal-2024-exploring-intra}, and \citet{musil-marecek-2024-exploring} who find interpretable axes in word embeddings using ICA; \citet{huertas-garcia-2023-dimensionality-reduction} who evaluate ICA as a dimensionality reduction technique for contextual sentence embeddings; \citet{aida-bollegala-2025-scdtour} who study semantic drift in time with ICA; and \citet{liu-2025-pruninglargelanguagemodels} use ICA to identify key neurons affecting LLM performance for pruning. Other notable recent work with LLM representations and ICA include \citet{liu-2025-pruninglargelanguagemodels} and \citet{abe-2025-latent-structures}. 

Another widespread tool to interpret structures in representational spaces is Sparse Auto-Encoders (SAEs, \citet{cunningham2023sparseautoencodershighlyinterpretable}). \citet{papadimitriou2025interpretinglinearstructure} explore the linear structures in the embedding spaces of vision-language models using SAEs, and find a balance between monomodal (i.e., either textual or visual) and modality-agnostic subspaces. \citet{saglam-etal-2025-large} compare the latent spaces of 11 causal language models and find that high-level semantic information exhibits low-dimensionality and linear separability. \citet{chen-etal-2025-knowledge} find SAEs to efficiently deconstruct polysemantic neurons, and \citet{lan2025quantifyingfeaturespaceuniversality} find universality in SAE-decompositions across different models.
Other ways to study structures include matrix based entropy (e.g., \citet{skean2025layer}), anisotropy (e.g., \citet{timkey-2021-all-bark}; \citet{mickus-etal-2022-dissecting-a-muppet}), singular value decomposition (e.g., \citet{chang-etal-2022-geometry}), clustering embeddings wrt. topics (e.g., \citet{teimouri-etal-2025-deep}), locally linear embeddings (e.g., \citet{lee2025sharedgeometry}), mapping hidden states with unembedding matrix (e.g., \citet{dar-etal-2023-analyzing}), and analyzing the effect of removing dimensions (\citet{takeshita-etal-2025-randomly}).

\section*{Limitations and Future Work}

The main limitation of our framework is the requirement of a one-to-one pairing of the datasets. This leaves multiclass classification and clustering out of our experiments, both of which are prominent embedding model tasks. Similarly, we also do not answer questions about the actual encoding of concepts themselves, rather the relationship they have with each other: our probes characterize \emph{geometric relationships} between paired representations (e.g., local neighborhood alignment and the concentration of query--target differences), rather than directly explaining how specific concepts are encoded. Embedding spaces likely contain additional non-linear or global phenomena that these probes cannot capture; this may already be reflected in tasks/datasets where correlations are weaker. We leave these relationships for future work.  Like any benchmark-driven analysis, our findings could be used for leaderboard-oriented optimization rather than general improvements.

Although we evaluate 25 models across five tasks and multiple languages, our findings may not generalize to all embedding models, datasets, or evaluation setups. Practical constraints (e.g., model access, the feasibility of running certain evaluations) limit coverage for some model--task combinations. Our analysis is also correlational: although our structural measures correlate with benchmark ranking, they do not establish which interventions would \textit{causally} improve a model. Future work could test causality by training or adapting models with explicit structure-preserving objectives and measuring downstream effects. Similarly, while all tested values of $k$ in $\text{Ret}_k$ supported our conclusions, understanding why certain values emerge as better estimators of performance is an open question we are eager to tackle. Our future work will focus on (i) extending the framework to non-paired tasks, (ii) deepening the analysis of prompting effects on structural alignment, (iii) incorporating non-linear probes (e.g., SAEs) to capture structure beyond ICA, and (iv) exploring how structure-based signals can support conditional or customizable embeddings, including properties that are not directly benchmarked.


\ifshowacks
    \acksection
    This research has been supported by Finland's Ministry of Education and Culture’s Doctoral Education Pilot under Decision No. VN/3137/2024-OKM-6 (The Finnish Doctoral Program Network in Artificial Intelligence, AI-DOC); the Digital Europe Programme under grant agreement No. 101195233 (OpenEuroLLM); the Human Diversity consortium, under the Profi7 program (grant no. 352727) and the Centre of Excellence (grant no. 374221) by the Research Council of Finland; the Research Council of Finland through FIN-CLARIAH research infrastructure (project 358720, which has also received funding from the European Union -- NextGenerationEU instrument), and the Mechanisms of Register Variation in Massively Multilingual Web-Scale Corpora (project 362459). This project is funded by the European Union. Views and opinions expressed are however those of the author(s) only and do not necessarily reflect those of the European Union or the European Education and Culture Executive Agency (EACEA). Neither the European Union nor EACEA can be held responsible for them.
We thank CSC -- IT Center for Science, Finland, for computational resources. 

\else
\fi


\bibliography{refs}


\newpage
\appendix
\section{Methodological Details}

In this section, we describe details relating to our methodology.

\subsection{Data Sampling}
\label{app:data-sampling}

We use the validation/development or test sets to ensure the models we use in the experiments are not trained on our data. We randomly sample up to 5000 documents from the datasets for all of our experiments to ensure manageable resource usage for our experiments: for example, the original WebFAQ dataset contains approximately 100k question-answer pairs for each language, hence using the full dataset would drastically increase the resources taken by embedding as well as time for fitting the ICA model. On the other hand, selecting 5000 examples is not possible in all cases; for instance, the SummEval dataset only contains 100 examples, of which we use 80 and reserve 20 as a test set for future work. 

Secondly, some of the datasets contain multiple types of relationships between the texts or multiple options for pair forming. SummEval dataset contains one document and multiple options for summary and associated quality scores, out of which we choose the one with the highest score. The availability of multiple summary options could facilitate using more than the best summary; however, this only increases the number of targets, not queries, and therefore, we opted to only include the best summaries. RTE3 datasets contain both entailing and contradicting passages, out of which we only consider the contradicting examples and discard the rest. We select the contradicting passages because using a mix of labels would be against our assumption of a clear, uniform signal across pairs, and using the entailing pairs would effectively turn RTE3-multi into an STS task.

Datasets are licensed under CC-BY-SA-4.0 (ARCChallenge), CC-BY-4.0 (WebFAQ, RTE3), MIT (SummEval), and CC-BY-2.0 (Tatoeba).

\subsection{Embedding models and Prompting}
\label{app:prompting}

\subsubsection{Prompting}
\begin{table}[!h]
    \centering
    \caption{Prompts used in our experiments.}
    \label{tab:prompts}
    \begin{tabular}{lc}
        \toprule
         Dataset & Prompt \\ \midrule
         ARCChallenge & \textit{Retrieve the answer to the question.}\\
         SummEval & \textit{Given a news article, retrieve semantically similar summaries.}\\
         RTE3-multi & \textit{Retrieve contradicting passages.}\\
         Tatoeba & \textit{Retrieve parallel sentences.} \\
         WebFAQ & \textit{Given a web search query, retrieve relevant passages that answer the query.}\\ \bottomrule
    \end{tabular}
\end{table}

Prompts used in our experiments are presented in Table \ref{tab:prompts}. The prompts were selected based on (1) MTEB-given prompt used in the evaluation, if available (ARCChallenge) (2) Prompts given by the dataset-developer, if available, or following examples of most typical prompts given by model developers or best practices in the field (others). While prompt sensitivity is a known feature of LLMs, the best choice would be to use a prompt optimized for each model and dataset separately. However, we leave these additional experiments for future work and therefore adopt the described approach. All prompts are inserted into the query with the template
\begin{tightquote}
    Instruct: \texttt{<prompt>} \\
    Query: \texttt{<query>}
\end{tightquote}
while the target is encoded as is.

\subsubsection{Embedding Model Licenses and Citations}

Below, we list the licenses for each embedding model we use in our experiments. 

\textbf{MIT}: BAAI/bge-m3 \citep{bge-m3}, BAAI/bge-base-en-v1.5 \citep{bge_v15}, intfloat/\-multiligual-e5-large-instruct \& intfloat/multilingual-e5-large \citep{wang2024multilingual-e5-large-instruct}, intfloat/e5-mistral-instruct \& intfloat/e5-small \citep{wang2022text-e5-instruct, wang2023improving-e5-instruct}, minishlab/potion-multilingual-128M \& minishlab/potion-base-8M \citep{minishlab2024model2vec}, HIT-TMG/KaLM-embedding-multilingual-mini-v1 \citep{hu2025kalmembeddingsuperiortrainingdata}, thenlper/gte-small \citep{li2023towards-gte-small}.

\textbf{Apache-2.0}: Jinaai/jina-b-en-v1 \citep{gunther2023jina}, Jinaai/jina-v2-small-en \citep{gunther2023jina2}, IBM-granite/granite-197-multilingual \& IBM-granite/granite-125M-english \citep{awasthy2025graniteembeddingmodels}, Google/LaBSE \citep{feng2022languageagnosticbertsentenceembedding-LaBSE}, Qwen/Qwen3-Embedding-8B \& Qwen/Qwen3-Embedding-0.6B \citep{qwen3embedding}, AliBaBaNLP/gte-multilingual-base \citep{zhang2024-gte}, Snowflake/snowflake-arctic-l-v2.0 \& Snowflake/snowflake-arctic-l \citep{yu2024arcticembed20multilingualretrieval}.

\textbf{CC-BY-4.0}: Google/gemini-embedding-001 \citep{geminiteam2025geminifamilyhighlycapable}, Google/EmbeddingGemma \citep{vera2025embeddinggemmapowerfullightweighttext}

\textbf{Other permissive}: NVIDIA/llama-embed-nemotron-8b \citep{babakhin2025llamaembednemotron8buniversaltextembedding}

\subsection{Independent Component Analysis}
\label{app:ICA}

In this section, we explain ICA and provide an illustrative example of its use. Similarly to PCA, the objective of ICA is to fit a matrix \mixer{}, known as \textit{unmixing matrix}, with a given set of constraints; instead of variance preservation, the target is the independence of resulting components \xica. Compared to PCA, which uses the covariance of the data to construct \mixer, ICA also includes higher order moments, skewness and kurtosis, to estimate the unmixing matrix \mixer{} \citep{yamagiwa-etal-2023-discovering}. 
The independent components produced by ICA can be calculated with
\[\xica = \embs \mixer,\]
$\mixer \in \Rnxn{d}{d_{ICA}}$ is the fitted unmixing matrix, $\embs \in \Rnxn{N}{d}$ is the observed signals, and $\xica \in \Rnxn{N}{d_{ICA}}$ is the resulting independent components.

The motivation behind ICA can be understood with a simple example: Assume $N$ microphones in a room with $M$ sound sources; each microphone catches the sound of multiple sources with different volumes. The recording of each microphone can be presented as a linear combination of all the sources $x_n = b_{1n}s_1 + b_{2n}s_2 + \dots + b_{nm}s_m $. Thus, the process of recoding the observed values $X$ can be represented as matrix multiplication
$X = SB.$
Furthermore, $X$ can be decoded into the sources $S$ again with the Moore-Penrose pseudo-inverse of $B^+ = \mixer$ as $S = X\mixer$. If $M = N$ and all sources are independent as assumed (i.e., $d = d_{ICA}$), the pseudo-inverse reduces to the regular matrix inverse. However, in a typical use case, only the observed matrix $X$ is known, and hence the matrix \mixer{} is estimated from observed values $X$ and the assumption of independence of sources. 

Using ICA on language model output embeddings may call into question the use of embeddings as the source signal: the most canonical examples of applying ICA are on time-series or otherwise \quot{continuous} data, like images, with embeddings lacking this structure. Despite this, ICA has been successfully applied to embeddings in previous work, as seen in our Background and Related Work sections. Another obstacle we face with ICA is the cardinality of our datasets. To fit an ICA model of dimension $\dim_{ICA} = N'$, we need at least $N'$ instances of embeddings; another reason we run our experiments with low-dimensional ICA.

\subsection{Computational Details}
\label{app:compute}

Our experiments cover 25 models, 29 datasets, and two prompting options (with and without prompt); altogether, we run experiments for 1450 combinations. Additionally, we run the MTEB evaluation for a subset of the combinations. We use AMD MI250x GPUs and AMD EPYC 7763 CPUs on \textit{an anonymous HPC environment} to run all experiments. All values reported below are top-end approximations.

On average, the experiments required 6-10 GB of memory, while some outliers, like the WebFAQ evaluation, required up to 20 GB. GPU usage varies based on the dataset size, both in time and in the number of GPUs used. We used 1-2 GPUs to embed the datasets and 2-8 GPUs to run MTEB evaluation. The total GPU computational cost of the embedding is  1650 GPU hours. The MTEB evaluation of ARCChallenge, Tatoeba, and SummEval required approximately 20 GPU hours, all with 2 GPUs, while the cost was significantly higher for WebFAQ, for which we estimate 1440 GPU hours, caused by the large size of the full evaluation set and the need to use 8 GPUs for some models. Calculating word embeddings for ICA explanations required approximately 10 GPU hours. 
All other experiments (calculation of neighborhood retention, fitting ICA models, calculating statistics, stability analysis) required a comparatively smaller amount of CPU hours: neighborhood retention experiment approximately 10 minutes/experiment, fitting ICA approximately 30 minutes (with some rare outliers taking more than 6 hours), and all analyses less than 5 minutes. In total, the computational cost for our experiment was 3120 GPU hours and 1100 CPU hours, or on average 2.15 hours of GPU and 45 minutes of CPU usage per dataset, model, and prompt combination.

Outside of the cost required to run the experiments in the paper, our preliminary experiments also required computational resources. For example, we experimented with values of $d_{\text{ICA}}$, which carries an associated CPU cost comparable to the CPU usage of the full experiment. However, as the embeddings could be saved and reused in all experiments, the preliminary experiments do not affect GPU hours substantially.

\section{Complementary Results}
\label{app:complementary_results}

In this section, we present the results for ICA$^{64}$, neighborhood retention results for suboptimal number of neighbors, analysis on the stability of ICA, and other supplementary examples and figures. 

\subsection{Results for ICA (dim=64)}
\label{app:ICA64}

Table \ref{tab:gini_ica64} presents the our results for ICA$^{64}$. Similar conclusions hold: ARCChallenge and Tatoeba receive high scores, while the scores of WebFAQ vary between languages. RTE3 dataset again shows increased scores compared to the neighborhood retention experiment. 

\begin{table}[]
    \centering
    \caption{Spearman correlation ($\uparrow$) for Gini-coefficient calculated from ICA$^{64}$ with MTEB scores, with bolded values for strong relationship ($\geq 0.7$). Star ($*$) indicates significance at level $p<0.05$, dagger ($\dagger$) at level $p<0.01$, and double dagger ($\ddagger$) at level $p<0.001$.}
    \label{tab:gini_ica64}
    \begin{tabular}{lllllll}
\toprule
Dataset         & Unpromted                   & Prompted                    &  & Dataset    & Unpromted          & Prompted                    \\ \cline{1-3} \cline{5-7} 
ARCChallenge  & \textbf{0.732}$^{\ddagger}$ & 0.669$^{\ddagger}$  & & SummEval  & 0.278 & 0.275 \\  \cline{1-3} \cline{5-7}
RTE3:eng  & 0.268 & 0.238  & & WebFAQ:bul  & -0.478$^{*}$ & -0.406 \\ 
RTE3:fra  & 0.608$^{\dagger}$ & 0.596$^{\dagger}$  & & WebFAQ:deu  & 0.595$^{\dagger}$ & \textbf{0.703}$^{\dagger}$ \\
RTE3:ita  & 0.561$^{\dagger}$ & 0.669$^{\ddagger}$  & & WebFAQ:ell  & -0.300 & -0.026 \\
RTE3:deu  & 0.417$^{*}$ & 0.499$^{*}$  & & WebFAQ:eng  & 0.290 & 0.447 \\ \cline{1-3}
Tatoeba:deu-eng  & \textbf{0.838}$^{\ddagger}$ & \textbf{0.945}$^{\ddagger}$  & & WebFAQ:fas  & -0.552$^{*}$ & -0.581$^{*}$ \\
Tatoeba:bul-eng  & \textbf{0.917}$^{\ddagger}$ & \textbf{0.895}$^{\ddagger}$  & & WebFAQ:fin  & \textbf{0.707}$^{\dagger}$ & 0.562$^{*}$ \\
Tatoeba:ukr-eng  & \textbf{0.937}$^{\ddagger}$ & \textbf{0.935}$^{\ddagger}$  & & WebFAQ:fra  & 0.659$^{\dagger}$ & 0.321 \\
Tatoeba:ita-eng  & \textbf{0.920}$^{\ddagger}$ & \textbf{0.898}$^{\ddagger}$  & & WebFAQ:hin  & -0.294 & -0.323 \\
Tatoeba:fra-eng  & \textbf{0.849}$^{\ddagger}$ & \textbf{0.879}$^{\ddagger}$  & & WebFAQ:ita  & \textbf{0.748}$^{\ddagger}$ & 0.688$^{\dagger}$ \\
Tatoeba:vie-eng  & \textbf{0.874}$^{\ddagger}$ & \textbf{0.884}$^{\ddagger}$  & & WebFAQ:jpn  & 0.659$^{\dagger}$ & 0.688$^{\dagger}$ \\
Tatoeba:ell-eng  & \textbf{0.872}$^{\ddagger}$ & \textbf{0.853}$^{\ddagger}$  & & WebFAQ:ukr  & -0.026 & -0.123 \\
Tatoeba:jpn-eng  & \textbf{0.854}$^{\ddagger}$ & \textbf{0.844}$^{\ddagger}$  & & WebFAQ:vie  & 0.490$^{*}$ & 0.614$^{\dagger}$ \\
Tatoeba:hin-eng  & \textbf{0.892}$^{\ddagger}$ & \textbf{0.885}$^{\ddagger}$  & & WebFAQ:zho  & 0.298 & 0.249 \\
Tatoeba:fin-eng  & \textbf{0.888}$^{\ddagger}$ & \textbf{0.868}$^{\ddagger}$  & & \\ \bottomrule
\end{tabular}
\end{table}

\subsection{Neighborhood Retention for Suboptimal Number of Neighbors}
\label{app:bad_k}

In the main body of the paper, we present the neighborhood retention experiment results for $k=10$, which was found to be in the top 3 values for all datasets. We remark that all tested values of $k$ support the conclusions. In Table \ref{tab:bad_neighbor_results} we justify this claim by presenting the results for one of the universally worst values of $k = 0.1N$ (and $k=5$ for SummEval), which was analogously found to be among the bottom 3 values. These results show that again, 4/5 datasets show significant strong to medium correlation, with RTE3 being the outlier. ARCChallenge shows noticeably smaller values for $k = 0.1N = 94$ than the optimal $k=10$ results, displaying that very local information between queries and targets has a lot of predictive power in this task. On the other hand, in Tatoeba and WebFAQ, some languages (like Greek and Finnish) show higher correlation when analyzed through the larger neighborhood ($k = 0.1N = 500$). We conclude that despite the differences, the overall conclusions we draw are still justified even with the worst value of $k$.

\begin{table}[]
    \centering
    \caption{Spearman correlation ($\uparrow$) for neighborhood retention with suboptimal $k=0.1N$ with MTEB scores, with bolded values for strong relationship ($\geq 0.7$). Star ($*$) indicates significance at level $p<0.05$, dagger ($\dagger$) at $p<0.01$, and double dagger ($\ddagger$) at $p<0.001$.}
    \label{tab:bad_neighbor_results}
\begin{tabular}{lccllcc}
\toprule
Dataset         & Unprompted                 & Prompted                   &  & Dataset    & Unprompted                 & Prompted                   \\ \cline{1-3} \cline{5-7} 
ARCChallenge  & \textbf{0.781}$^{\ddagger}$ & \textbf{0.866}$^{\ddagger}$  & & SummEval  & 0.587$^{\dagger}$ & 0.615$^{\dagger}$ \\ \cline{1-3} \cline{5-7} 
RTE3:eng  & -0.304 & -0.048  & & WebFAQ:bul  & \textbf{0.903}$^{\ddagger}$ & \textbf{0.915}$^{\ddagger}$ \\
RTE3:fra  & 0.149 & 0.268  & & WebFAQ:deu  & \textbf{0.843}$^{\ddagger}$ & \textbf{0.841}$^{\ddagger}$ \\
RTE3:ita  & 0.344 & 0.495$^{*}$  & & WebFAQ:ell  & \textbf{0.977}$^{\ddagger}$ & \textbf{0.915}$^{\ddagger}$ \\
RTE3:deu  & 0.009 & 0.180  & & WebFAQ:eng  & 0.538$^{*}$ & 0.453 \\ \cline{1-3} 
Tatoeba:deu-eng  & \textbf{0.853}$^{\ddagger}$ & \textbf{0.798}$^{\ddagger}$  & & WebFAQ:fas  & \textbf{0.940}$^{\ddagger}$ & \textbf{0.882}$^{\ddagger}$ \\
Tatoeba:bul-eng  & \textbf{0.964}$^{\ddagger}$ & \textbf{0.897}$^{\ddagger}$  & & WebFAQ:fin  & \textbf{0.946}$^{\ddagger}$ & \textbf{0.872}$^{\ddagger}$ \\
Tatoeba:ukr-eng  & \textbf{0.922}$^{\ddagger}$ & \textbf{0.876}$^{\ddagger}$  & & WebFAQ:fra  & \textbf{0.891}$^{\ddagger}$ & \textbf{0.831}$^{\ddagger}$ \\
Tatoeba:ita-eng  & \textbf{0.916}$^{\ddagger}$ & \textbf{0.852}$^{\ddagger}$  & & WebFAQ:hin  & \textbf{0.880}$^{\ddagger}$ & \textbf{0.847}$^{\ddagger}$ \\
Tatoeba:fra-eng  & \textbf{0.878}$^{\ddagger}$ & \textbf{0.829}$^{\ddagger}$  & & WebFAQ:ita  & \textbf{0.893}$^{\ddagger}$ & \textbf{0.775}$^{\ddagger}$ \\
Tatoeba:vie-eng  & \textbf{0.953}$^{\ddagger}$ & \textbf{0.870}$^{\ddagger}$  & & WebFAQ:jpn  & \textbf{0.874}$^{\ddagger}$ & \textbf{0.837}$^{\ddagger}$ \\
Tatoeba:ell-eng  & \textbf{0.962}$^{\ddagger}$ & \textbf{0.875}$^{\ddagger}$  & & WebFAQ:ukr  & \textbf{0.940}$^{\ddagger}$ & \textbf{0.862}$^{\ddagger}$ \\
Tatoeba:jpn-eng  & \textbf{0.952}$^{\ddagger}$ & \textbf{0.888}$^{\ddagger}$  & & WebFAQ:vie  & \textbf{0.891}$^{\ddagger}$ & \textbf{0.812}$^{\ddagger}$ \\
Tatoeba:hin-eng  & \textbf{0.930}$^{\ddagger}$ & \textbf{0.810}$^{\ddagger}$  & & WebFAQ:zho  & \textbf{0.841}$^{\ddagger}$ & \textbf{0.849}$^{\ddagger}$ \\
Tatoeba:fin-eng  & \textbf{0.975}$^{\ddagger}$ & \textbf{0.914}$^{\ddagger}$  & &\\ \bottomrule
\end{tabular}
\end{table}

\subsection{Stability of ICA}
\label{sec:stability}
As the FastICA algorithm is non-deterministic, analysis of the stability of the results we acquire from these methods is necessary. To evaluate this, we use 4 models (LaBSE, Qwen3-Embedding-0.6B, bge-m3, multilingual-e5-large). We opted to use Tatoeba and these models for this analysis, as analyzing the stability of the peak requires the presence of a sustained peak in the results. We fit 8 differently initialized ICA$^{32}$ models per dataset-model combination and evaluate the stability of the Gini-coefficient measure and the peak-related direction extracted by ICA. In this experiment, we use only the unprompted embeddings to limit the number of models needed to fit in this step and because we are evaluating the stability of our methodology, not the effect of prompts.

We first evaluate the stability of the Gini coefficient. We find that among the different initializations, the Gini coefficient is stable, measured with the strict coefficient of variation at threshold 0.01, in all model-dataset combinations except for Qwen-embedding-0.6B and Finnish (and even in this case, the standard deviation across runs is just above the threshold). This overwhelming stability led to our usage of Gini over other metrics (peak-to-mean-ratio, coefficient of variation).

The other question on the stability of ICA is how consistent the fitted model is. The Icasso method \citep{Himberg-2004-Icasso} has been suggested to evaluate the fluctuations of results produced by ICA.
We modify this setup slightly and incorporate greedy dimension matching inspired by the work of \citet{yamagiwa-etal-2023-discovering}, as we are mostly interested in the stability of the peak dimension, which we assign with $\arg\max$-function for this step. Following \citet{li-etal-2024-exploring-intra}, we measure stability with 
\begin{align*}
    \label{alignment}
    \sigma_{ij} = \absvalue{
    \frac{
        \text{Cov}(s_i, s'_j)
        }{
        \sqrt{\frac{1}{d} \sum_k s_{ik}^2} \sqrt{ \frac{1}{d} \sum_k {s'}_{jk}^2}
        }
    },
\end{align*}
where $s_i$ and $s'_j$ are rows from the unmixing matrices from different runs. 

\begin{figure}%
    \centering
    \subfloat[\centering Cross-run stability of ICA]{{\includegraphics[width=0.40\textwidth]{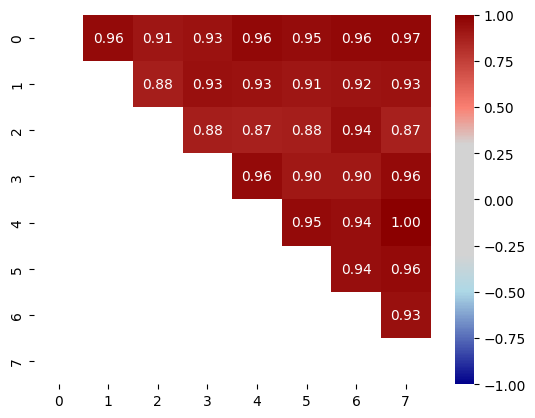} }}%
    \qquad
    \subfloat[\centering Cross-run stability of ICA peaks]{{\includegraphics[width=0.40\textwidth]{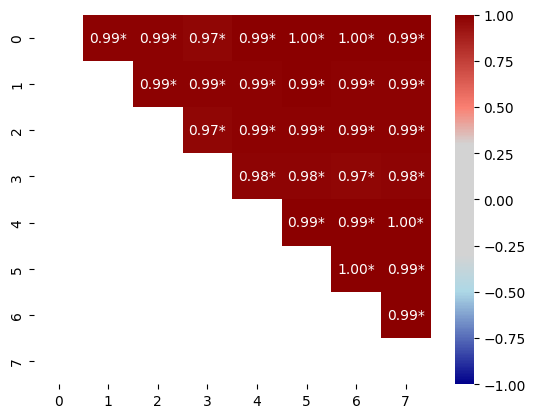} }}%
    \caption{(a) Aggregated similarity between the components of 8 differently initialized ICA models (b) Similarity between peak components across runs. Both heatmaps are on LaBSE and Tatoeba:eng-deu. Star (*) indicates significance at level $p<0.05$. }%
    \label{fig:stability_example}%
\end{figure}

Our results also show that ICA can reliably produce equivalent results: the row values of the unmixing matrices corresponding to the peak correlate strongly across runs, meaning ICA is able to latch onto the \textit{same representation in the embeddings}, an example of which is shown in Figure \ref{fig:stability_example}. Out of the ten languages tested, LaBSE and multilingual-e5-large display stability comparable to the exemplified figures. For Qwen3-Embedding-0.6B and bge-m3, we reach similarly comparable stability for 9/10 and 7/10 languages. Even in the negative cases, most comparisons reach correlations comparable to the example: For example, on bge-m3 and the Ukrainian split, 6/8 ICA models from the 8 runs agree on the peak, but the remaining 2/8 only correlate with each other. We conclude that, on average, we reach high stability on the fitted ICA models. 

\begin{figure}%
    \centering
    \subfloat[\centering text-embedding-3-small, Tatoeba:fin-eng]{
        {\includegraphics[width=0.5\textwidth]{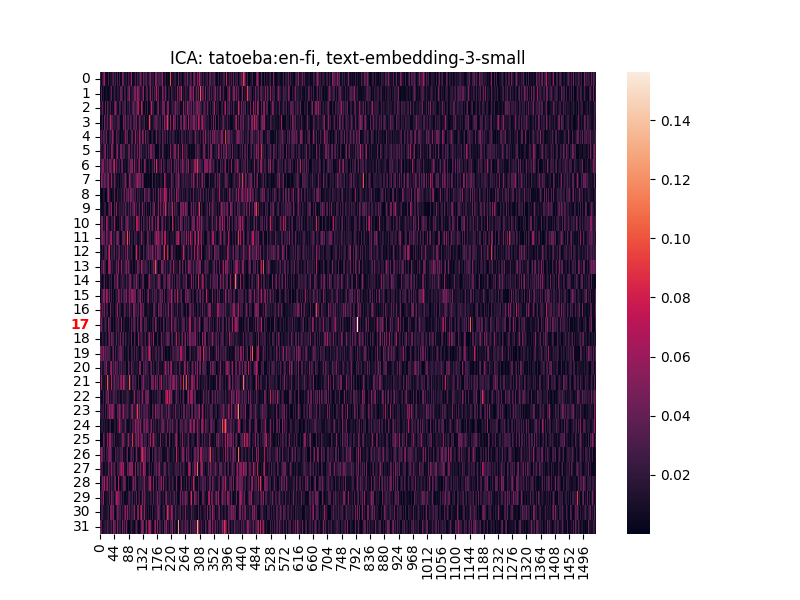} }
    }%
    \subfloat[\centering llama-embed-nemotron-8b, WebFAQ:jpn]{
        {\includegraphics[width=0.5\textwidth]{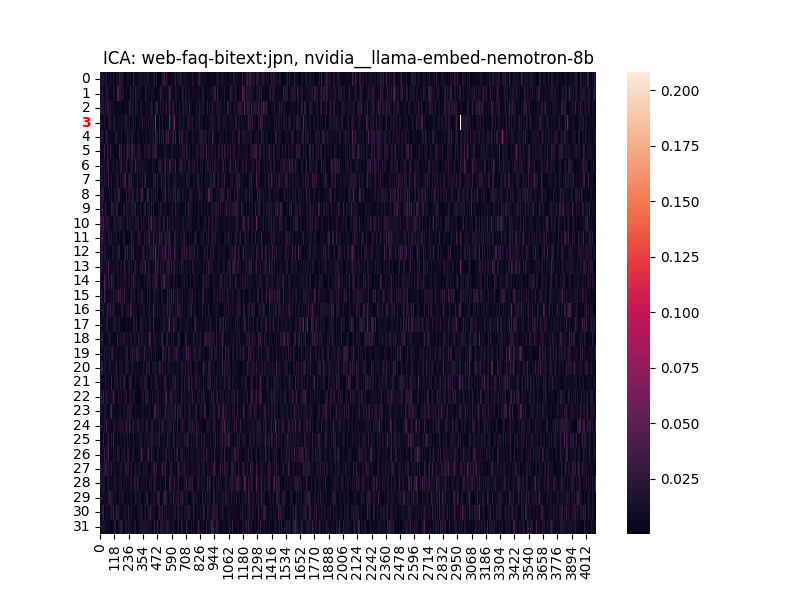} }
    } 
    \caption{Two unmixing matrices displaying one or two embedding model dimensions that strongly contribute to the peak dimension (bolded and red) in ICA. }%
    \label{fig:unmixers}%
\end{figure}

\subsection{Interpretability through ICA}
\label{sec:keywords+matrix}
We follow \citet{li-etal-2024-exploring-intra} and \citet{yamagiwa-etal-2024-axistour} and explain the results of the fitted ICA models by using word embeddings of the top 10k words per language \citep{top10000words-distributor}, and select the words that obtain the highest and lowest values for the peak dimension. Again, we use only the unprompted embeddings (the prompt would likely affect the keywords for each peak). For Tatoeba, we use both English and non-English word lists, and for WebFAQ and RTE-multi, we use the appropriate languages. These results allow us to see which words each ICA dimension corresponds to the strongest. 

Examples of results are given in Table \ref{tab:word-explanations}. From the selection of words, it is clear which side corresponds to the query and which to the target. Although not perfect, the presence of Finnish translation-related vocabulary on the Tatoeba:eng-fin task, and the American and French related vocabulary on Tatoeba:eng-fra task, suggests that the peak seen in our experiments does correspond to a language-specific signal. Similarly, in the ARCChallenge task, the top 10 words are associated with answers (animals, shapes, and units), as the bottom words can be seen as educational, opinionated, and questioning, corresponding to the contents of the ARCChallenge dataset. We concede that the perceived quality for the obtained word explanations is heavily model and dataset-dependent; for example, SummEval or RTE3 do not produce a clearly categorizable set of words.

Another way of using ICA for interpretation is to investigate the fitted unmixing matrix. Figure \ref{fig:unmixers} presents the matrix for llama-embed-nemotron-8b on Tatoeba:eng-fin task and llama-embed-nemotron-8b on WebFAQ:jpn, with vertical values indicating dimensions of the ICA model and horizontal values corresponding to the dimensions of the embedding model. In the first example, the dimension corresponding to the peak, 17, shows one distinctive high value, highlighting the dimension of the original model that most corresponds to the peak. The same can be seen on the second example, where two dimensions are highlighted as most contributing to the peak. This means that Finnish translation and Japanese question-answering rely on only one or two dimensions in the respective models.

\begin{table}[]
    \small
    \centering
    \caption{Examples of top and bottom 10 most salient words for the peak-dimension in ICA$^{32}$, with translations. The emergence of \quot{Finnish translation}, \quot{American}, \quot{Notre-Dame} clearly indicates the presence of translation, language, and location-related signals, while units, shapes, and animals indicate common answers to questions in the ARCChallenge dataset.}
    \label{tab:word-explanations}
    \renewcommand\arraystretch{1.2}
        \begin{tabular}{p{3.7cm}p{3.7cm}p{5.5cm}}
        \toprule
        \textbf{Model \& Dataset} & \textbf{Peak top 10} & \textbf{Peak bottom 10} \\
        \midrule
        \makecell[l]{llama-embed-nemotron-8b \\Tatoeba:eng-fin}
        &
        \makecell[l]{quotations, quotes, quote,\\ maybe, perhaps, memories, \\  somehow, remember,\\  sometimes, remembered}
        &
        \makecell[l]{
            \transl{do-you-use,}{käytätkö} 
            \transl{use (impr.),}{käyttäkää} \\
            \transl{translated-to-Finnish,}{suomentanut} \\
            \transl{Finnish-translation,}{suomennos}\\
            \transl{I-send,}{lähtetän} 
            \transl{can-you,}{voitteko} \\
            \transl{could-you,}{voisitteko} 
            \transl{do-you-want,}{tahdotko}\\
            \transl{would-you-take,}{ottaisitko}
            \transl{you-use,}{käytät}
        }
        \\
        \midrule
        \makecell[l]{gte-multilingual-base \\Tatoeba:eng-fra}
        &
        \makecell[l]{travelling, drinking, USA, \\ World, living, eating, \\ speech, American,  \\ America, speaking}
        &
        \makecell[l]{
            \transl{I love,}{J'aime} 
            \transl{calculated,}{calculée} \\
            Notre-Dame,
            \transl{It was,}{C'était}
            \transl{it was,}{c'était} \\
            \transl{calculated,}{calculé}
            \transl{I,}{Je} 
            \transl{time,}{l'heure}\\ 
            \transl{Yet,}{Pourtant}
            \transl{love,}{l'amour}
        }
        \\
        \midrule
        \makecell[l]{e5-mistral-instruct \\ARCChallenge}
        &
        \makecell[l]{rope, worm, needle, meters, \\lightning, dried, snake, \\cylinder, meters, frog}
        &
        \makecell[l]{education, opinions, justice,  educators, \\stakeholders,  politics, policies, \\ attitudes, teachers
        }
        \\
        \bottomrule
        \end{tabular}
\end{table}

\subsection{Complementary figures}
\label{app:complementary_figures}

\begin{figure}[b]
    \scriptsize
    \centering
    \includegraphics[width=0.85\linewidth]{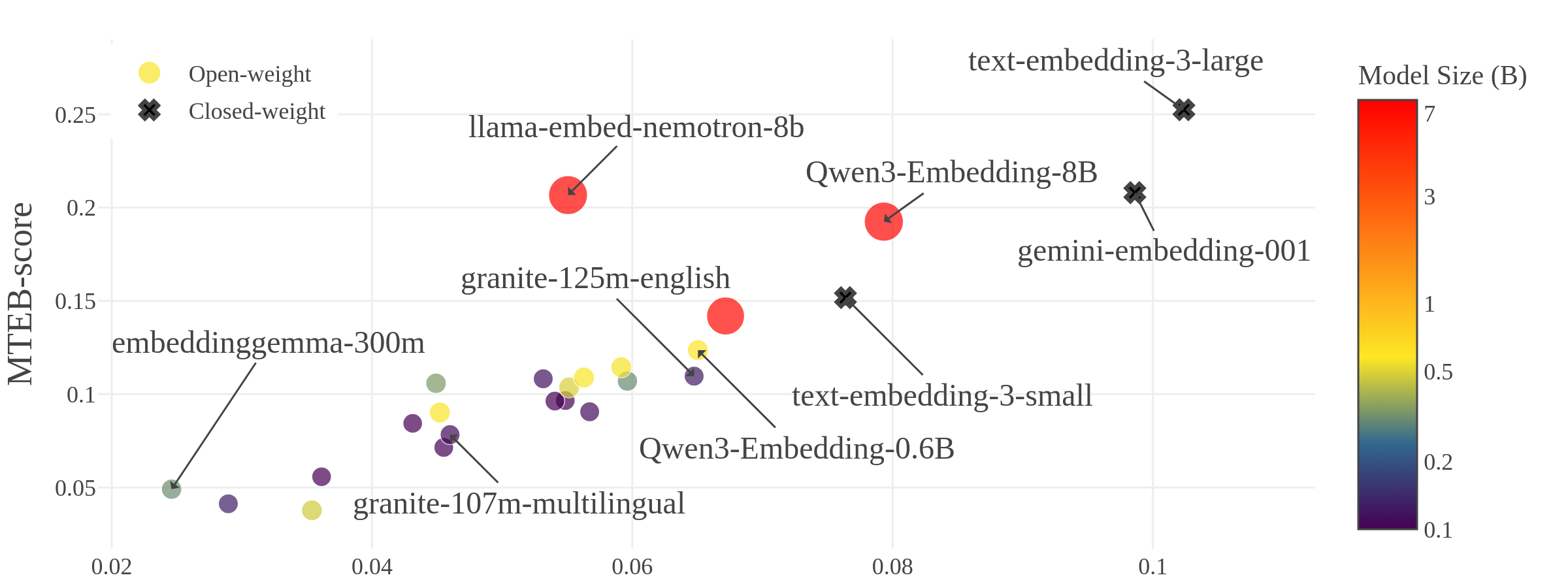}
    \caption{ARCChallenge: Neighborhood retention (horizontal) vs. MTEB-score (Spearman $r$ \textbf{0.890}$^{\ddagger}$). This figure shows that outside the clear outlier, llama-embed-nemotron-8b, the relationship between neighborhood retention and MTEB-score is strong. The largest models receive both the highest MTEB performance and retention scores. Among the smaller models, size shows more variation.}
    \label{fig:knn:ARRChallenge}
\end{figure}

\begin{figure}
    \scriptsize
    \centering
    \includegraphics[width=0.85\linewidth]{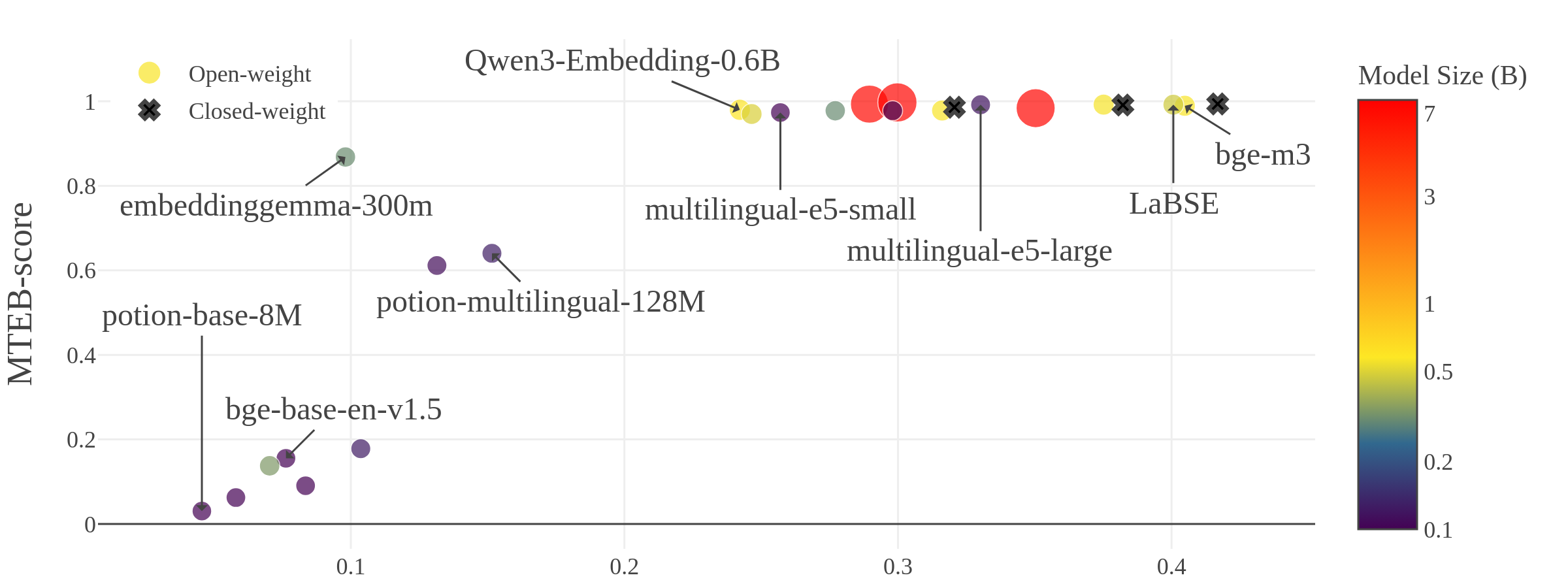}
    \caption{Tatoeba:deu-eng: Neighborhood retention (horizontal) vs. MTEB-score (Spearman $r$ \textbf{0.888}$^{\ddagger}$). German is among the least correlated Tatoeba languages in our experiment. Clearly, one possible cause for this is the fact that so many models receive close to perfect scores on this task, reflecting the overall good German capabilities of the tested models. Slight outlier behavior can be seen from embeddinggemma-300m. Otherwise, similar conclusions hold as for the other Tatoeba tasks, namely that multilingual versions outperform the monolingual versions within model families.}
    \label{fig:knn:Tatoeba:deu-eng}
\end{figure}

\begin{figure}
    \scriptsize
    \centering
    \includegraphics[width=0.85\linewidth]{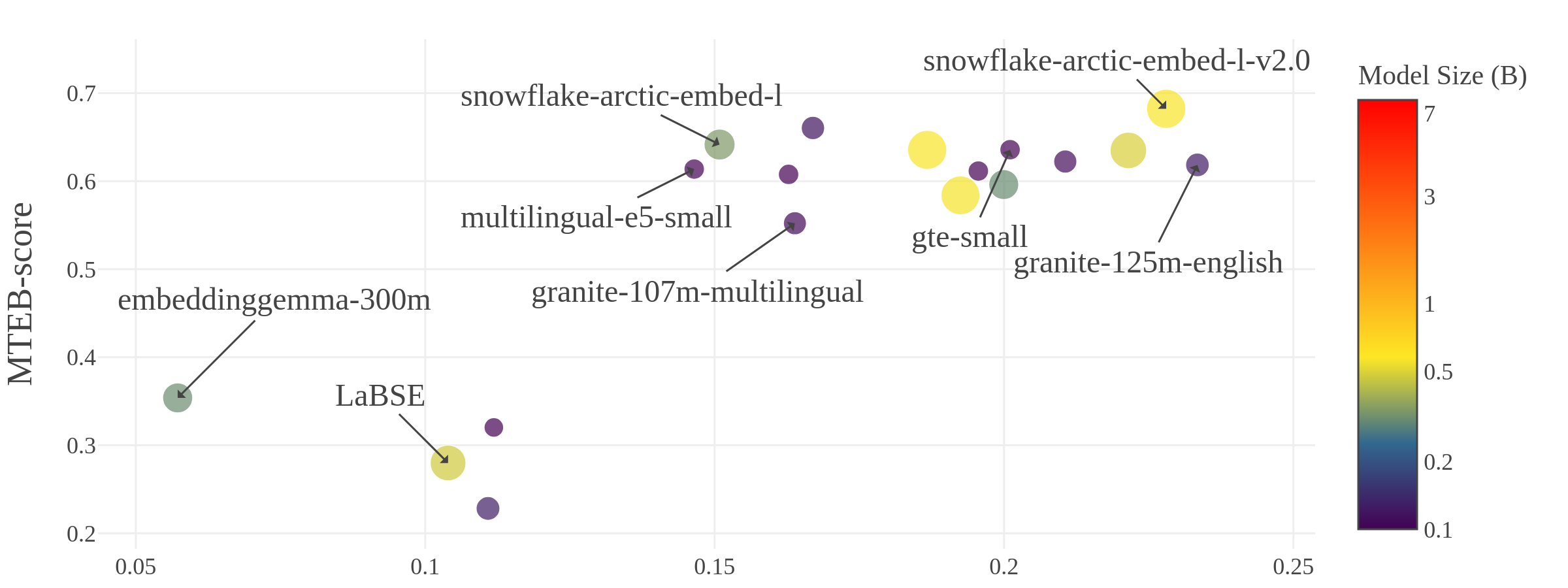}
    \caption{WebFAQ:eng: Neighborhood retention (horizontal) vs. MTEB-score (Spearman $r$ 0.620$^{\dagger}$). The English subset of WebFAQ was a clear outlier in our results, with a low correlation coefficient. Compared to Figure \ref{fig:knn:WebFAQ:ell}, which shows Greek results (among the best), it is clear where the difference emerges: on the English WebFAQ task, almost all models reach similar performance, and only 4 models struggle on the task. This is in line with the fact that English is the most used language for language model training, and leads to a situation where, unlike with other languages, models with middle range scores of 0.4--0.5 do not exist, and therefore they are arranged into two clusters--those that solve the task and those that struggle--instead of the relationship seen in the other languages.}
    \label{fig:knn:WebFAQ:eng}
\end{figure}

\begin{figure}
    \scriptsize
    \centering
    \includegraphics[width=0.85\linewidth]{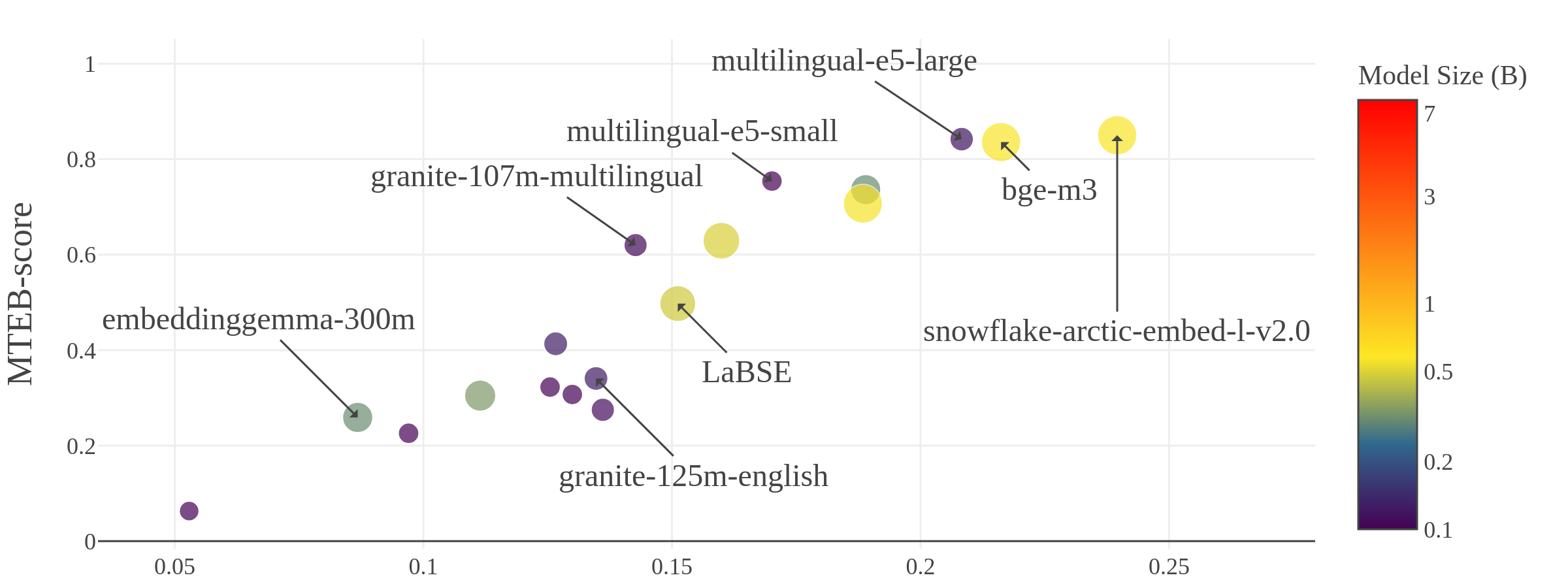}
    \caption{WebFAQ:ell: Neighborhood retention (horizontal) vs. MTEB-score (Spearman $r$ \textbf{0.946}$^{\ddagger}$). This Greek dataset received one of the highest correlation scores in WebFAQ experiments. Compare to Figure \ref{fig:knn:WebFAQ:eng}. }
    \label{fig:knn:WebFAQ:ell}
\end{figure}

\begin{figure}
    \scriptsize
    \centering
    \includegraphics[width=0.85\linewidth]{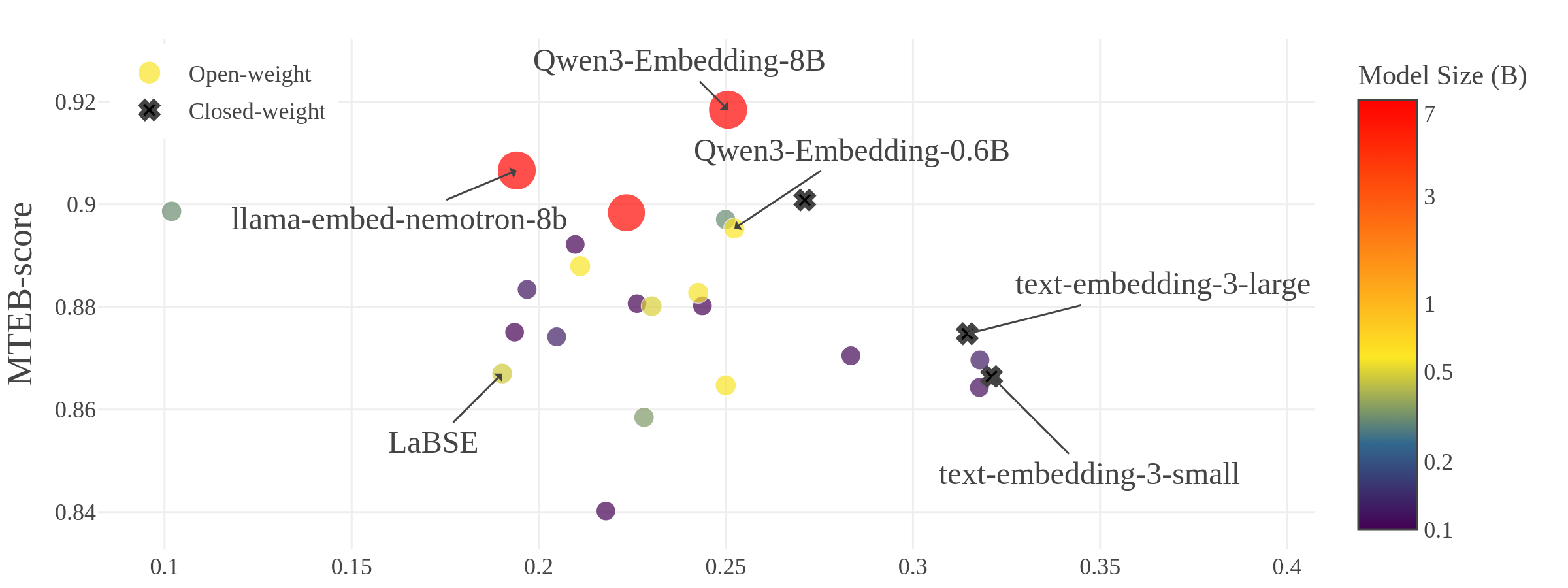}
    \caption{RTE3:eng: Neighborhood retention (horizontal) vs. MTEB-score (Spearman $r$ $-0.230$). As seen in Section \ref{sec:results}, RTE3-multi shows almost no relationship between retention and evaluation score, which we tied to the way MTEB evaluates contradiction in Section \ref{sec:results}: instances close to each other are classified as entailing, while instances far apart are classified as contradicting.}
    \label{fig:knn:RTE-en}
\end{figure}

\begin{figure}
    \centering
    \scriptsize
    \includegraphics[width=0.85\linewidth]{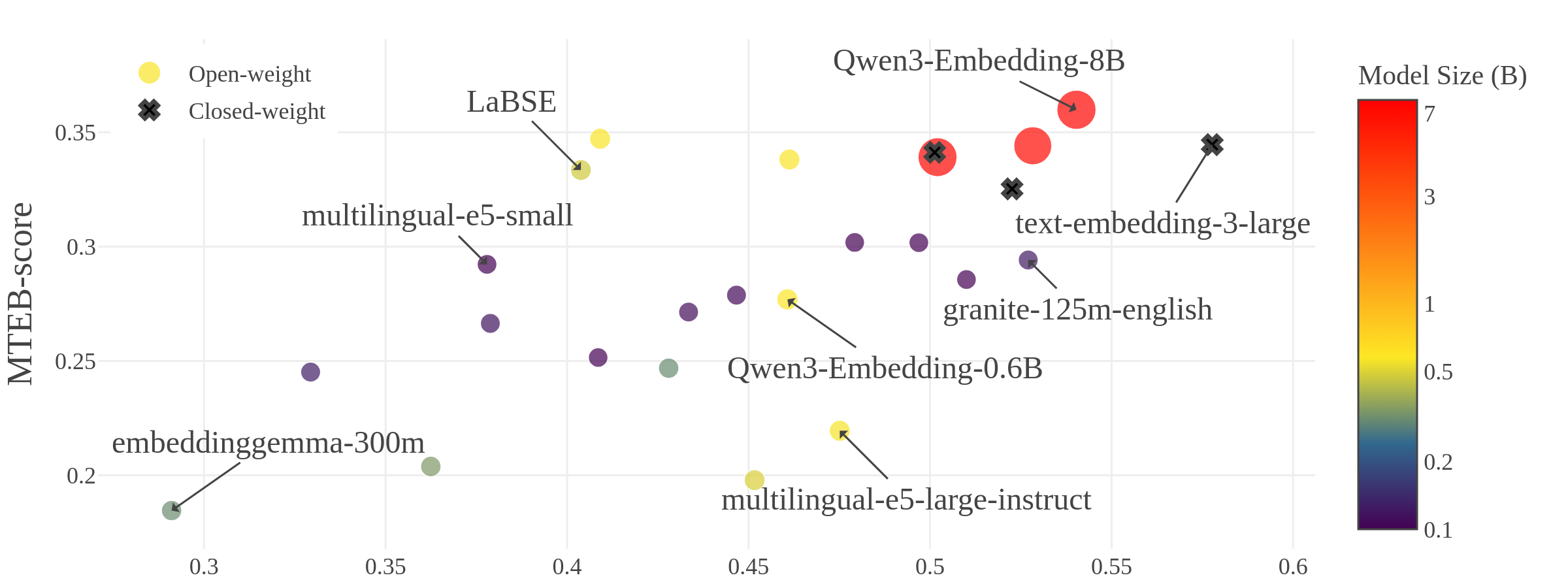}
    \caption{SummEval: Neighborhood retention (horizontal) vs. MTEB-score (Spearman $r$ 0.643$^{\ddagger}$). These results display a lot of variance in the middle range of retention, while the largest models are clustered closely together at the highest end. Small models, like granite-125M-english, reach comparable results and retention to the largest models.}
    \label{fig:knn:Summeval}
\end{figure}

\begin{figure}[!hb]%
    \centering
    \subfloat[\centering ARCChallenge (\textcolor{brickred}{$\bullet$} 0.810$^\ddagger$ \textcolor{navyblue}{$\blacklozenge$} 0.689$^\ddagger$)]{{\includegraphics[width=6cm]{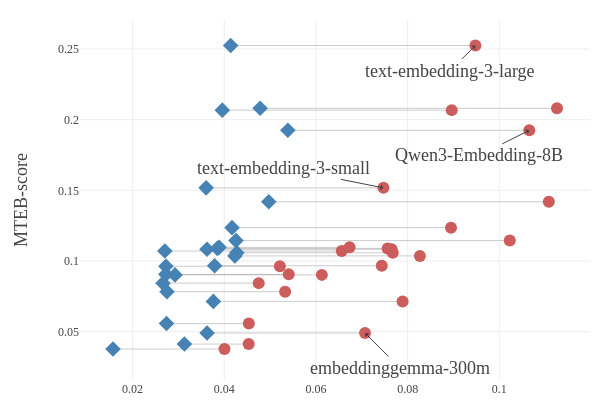} }}%
    \qquad
    \subfloat[\centering SummEval (with prompt, \textcolor{brickred}{$\bullet$} 0.428$^*$, \textcolor{navyblue}{$\blacklozenge$} 0.220 )]{{\includegraphics[width=6cm]{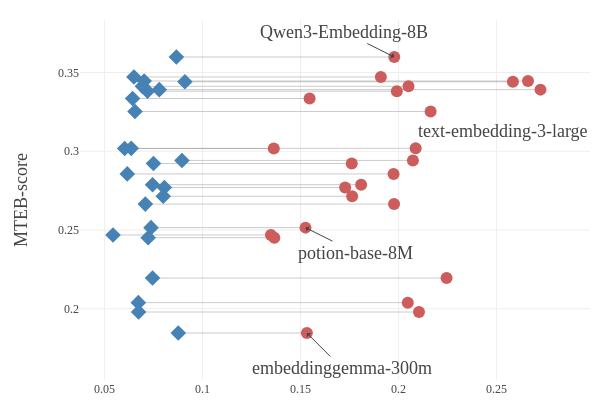} }}%
    \caption{Shuffling experiment visualisation for two additional datasets. (a) Shuffling affects ARCChallenge, but the relationship is still noticeable and significant after shuffling. This means that while neighborhood retention correlated even more strongly with performance on ARCChallenge, global information is still enough to distinguish model performance differences. (b) The effect on prompted SummEval: Shuffling affects the correlation heavily. SummEval saw medium correlation in neighborhood retention, which leads us to conclude that summarization is encoded non-linearly and locally on average in the models we test. }%
    \label{fig:shuffling:ARC+SummEval}%
\end{figure}

In this section, we present supplementary figures. Figures \ref{fig:knn:ARRChallenge}--\ref{fig:knn:Summeval} display visualizations of MTEB-scores against $\text{Ret}_k$ with annotated model names and parameter counts, and Figure \ref{fig:shuffling:ARC+SummEval} presents shuffling effect visualization for ARCChallenge and SummEval. Each caption contains a short analysis of the most important takeaways.

    %




\end{document}